\documentclass[journal]{IEEEtran}

\IEEEoverridecommandlockouts

\usepackage{graphics} % for pdf, bitmapped graphics files
\usepackage{epsfig} % for postscript graphics files
\usepackage{mathptmx} % assumes new font selection scheme installed
\usepackage{times} % assumes new font selection scheme installed
\usepackage{amsmath} % assumes amsmath package installed
\usepackage{amssymb}  % assumes amsmath package installed
\usepackage{multirow}
\usepackage[ruled]{algorithm2e}
\usepackage{epstopdf} %transfer eps to pdf
\usepackage{subfigure} %for subfigures
\usepackage{stfloats} %for table
\usepackage{bm}
\usepackage{color}
\usepackage{amsthm}
\usepackage{cite}
\usepackage{array}
\newtheorem{remark}{Remark}
\usepackage{threeparttable}
\usepackage{nomencl}
\usepackage{braket}
\usepackage[switch,pagewise,columnwise]{lineno}

\makenomenclature

\begin{document}

\title{Deep Reinforcement Learning with Quantum-inspired Experience Replay}
\author{Qing~Wei, Hailan~Ma, Chunlin~Chen,~\IEEEmembership{Member,~IEEE}, Daoyi~Dong,~\IEEEmembership{Senior Member,~IEEE}
\thanks{This work was supported in part by the National Natural Science Foundation of China (No.71732003 and No.61828303) and by the Australian Research Council's Discovery Projects funding scheme under project DP190101566.}

\thanks{Q. Wei, H. Ma and C. Chen are with the Department of Control and Systems Engineering, School of Management and Engineering, Nanjing
University, Nanjing 210093, China (e-mail: clchen@nju.edu.cn).}
\thanks{D. Dong is with the School of Engineering and Information Technology,
University of New South Wales, Canberra, ACT 2600, Australia (email: daoyidong@gmail.com).}}

\maketitle
\begin{abstract}
In this paper, a novel training paradigm inspired by quantum computation is proposed for deep reinforcement learning (DRL) with experience replay. In contrast to traditional experience replay mechanism in DRL, the proposed deep reinforcement learning with quantum-inspired experience replay (DRL-QER) adaptively chooses experiences from the replay buffer according to the complexity and the replayed times of each experience (also called transition), to achieve a balance between exploration and exploitation. In DRL-QER, transitions are first formulated in quantum representations, and then the preparation operation and the depreciation operation are performed on the transitions. In this progress, the preparation operation reflects the relationship between the temporal difference errors (TD-errors) and the importance of the experiences, while the depreciation operation
 is taken into account to ensure the diversity of the transitions. The experimental results on Atari 2600 games show
that DRL-QER outperforms state-of-the-art algorithms such as DRL-PER and DCRL on most of these games with improved training efficiency, and is also applicable to such memory-based DRL approaches as double network and dueling network.
\end{abstract}

\begin{IEEEkeywords}
 Deep reinforcement learning, quantum reinforcement learning, quantum computation, quantum-inspired experience replay.
\end{IEEEkeywords}

\nomenclature{$S$}{State space.}%

\nomenclature{$\epsilon$}{$\epsilon$-greedy factor.}%

\nomenclature{$A$}{Action space.}%

\nomenclature{$P$}{State transition probability.}%

\nomenclature{$R$}{Reward function.}%

\nomenclature{$s_t$}{State at time step $t$.}%

\nomenclature{$a_t$}{Selected action according to policy $\pi(s_t)$ at time $t$.}%

\nomenclature{$T$}{Terminal time of one episode.}%

\nomenclature{$\pi$}{Policy function.}%

\nomenclature{$r_t$}{Scalar reward at time step $t$.}%

\nomenclature{$\gamma$}{Discount factor.}%

\nomenclature{$Q(s,a)$}{State-action values.}%

\nomenclature{$\theta$}{Parameters of the evaluation network.}%

\nomenclature{$\theta^-$}{Parameters of the target network.}%

\nomenclature{$k$}{Label of experience in buffer.}%

\nomenclature{$\delta$}{Temporal-difference error.}%

\nomenclature{$e_t$}{The $t$-th experience.}%

%\nomenclature{$b_k$}{the significance of the $e_k$.}%

\nomenclature{$\alpa$}{Complex coefficient of $|0\rangle$.}%

\nomenclature{$\beta$}{Complex coefficient of $|1\rangle$.}%

%\nomenclature{$|\psi'$}{The state vector after evolution.}%

\nomenclature{$U$}{Unitary operator.}%

\nomenclature{$X_h$}{Projector onto the eigenspace of $H$ with $h$.}%

\nomenclature{$h$}{Eigenvalue.}%

\nomenclature{$P(h)$}{Probability of obtaining $h$.}%

\nomenclature{$b_0^{(k)}$}{Coefficient of $|0\rangle$ of $|\psi^{(k)}\rangle$.}%

\nomenclature{$b_1^{(k)}$}{Coefficient of $|1\rangle$ of $|\psi^{(k)}\rangle$.}%

\nomenclature{$\Phi$}{Rotation angle.}%

\nomenclature{$Y$}{Pauli $Y$ matrix.}%

\nomenclature{$\rm{i}$}{$\rm{i}=\sqrt{-1}$.}%

\nomenclature{$\sigma$}{Preparation factor.}%

\nomenclature{$\omega$}{Depreciation factor.}%

%\nomenclature{$\Sigma$}{$\Sigma=m\sigma$.}%

\nomenclature{$\ket{\psi}$}{State vector (quantum state).}%

\nomenclature{$RT_{\max}$}{Maximum value of times of being replayed.}%

\nomenclature{$\delta_{\tilde{i}, i}$}{Kronecker delta.}%

\nomenclature{$\mathcal{H}$}{Hilbert space.}%

\nomenclature{$U^{\dagger}$}{Transposed conjugate matrix of $U$.}%

\nomenclature{$\bigotimes$}{Tensor product.}%

\nomenclature{$\mu$}{Hyper-parameters for the rotation time $m_k$.}%

\nomenclature{$\iota$}{The angle of the uniform state.}

\nomenclature{$\zeta_{1,2}$}{Hyper-parameters for the preparation factor  $\delta$.}%

\nomenclature{$\tau_{1,2}$}{Hyper-parameters for the depreciation factor $\omega$.}%

\nomenclature{$M$}{Size of experience replay buffer.}%

\nomenclature{$SK$}{Label of the selected experience in experience replay.}%

%makeindex DRL-QER.nlo -s nomencl.ist -o DRL-QER.nls
\printnomenclature

\section{Introduction}\label{Sec1}
\IEEEPARstart{R}{einforcement} learning (RL) is an intelligent paradigm that learns
through the interaction with the environment.
During the training process of this interaction-based algorithm,
similar to human behaviours \cite{sutton2018reinforcement,Littman2015,li2020quantum}, the
agent adjusts its behavior to maximize the cumulative rewards for the entire control task according to the retroaction it receives from the
environment.
When it comes to the general situation of the real world
environment, most control tasks often come with high-dimensional inputs,
where traditional RL approaches cannot work well.
Fortunately, deep learning provides a new approach to handle the complex input information and has achieved huge breakthrough in various fields \cite{krizhevsky2012imagenet,farabet2012learning,goh2014learning,hinton2012deep,sutskever2014sequence,brahma2015deep,Tembine2020}. In particular, by combining deep learning with RL, a new framework of deep reinforcement learning (DRL) arises, where Deep Q 
Network (DQN) becomes one of the most famous DRL methods \cite{mnih2015human}.

DQN was employed to estimate the action values to help the agent make sequential decisions, where raw images were fed into the convolutional neural networks followed by the fully connected networks to output the action values that estimated the future rewards. In order to improve the utilization of the state-action transitions,
an experience replay mechanism was deployed in the DQN framework \cite{lin1992self}, where experiences were stored in a finite-size buffer and were retrieved from the buffer. This mechanism of experience replay effectively sped up the processing of the experiences during training, but it ignored the difference in the importance between experiences. A series of experience replay variants
have been developed to further improve the learning process,
such as  prioritized experience replay (PER) \cite{schaul2015prioritized}, deep curriculum reinforcement learning (DCRL) \cite{ren2018}, remember and forget for experience replay (ReF-ER) \cite{pmlr-v97-novati19a}, attentive experience replay (AER) \cite{SunZL20} and
competitive experience replay (CER) \cite{LiuTSX19}.

In PER, temporal difference errors (TD-errors) determine the priorities of the experiences and influence the probabilities of those experiences' being replayed. 
This method further enhance the utilization of experiences compared to the original experience replay, but the data generated by the agent are noisy. 
Since DRL-PER gives higher priorities to transitions with larger TD errors and there might exist some experiences whose large TD-errors would not decrease even after many times of replay. 
From this respective, DRL-PER may cause some experiences to be overused, which might result in oscillations of the neural network \cite{chang2017active}. To improve the sample efficiency of DQN, DCRL proposed a criterion for the samples' importance based on the difficulties and the diversities of the experiences, where the difficulties are positively correlated with TD-errors and  
the diversities are related to the number of replaying times \cite{ren2018}. However, it introduced a number of parameters that required more prior knowledge to tune accurately. In ReF-ER \cite{pmlr-v97-novati19a}, policy updates are penalized according to Kullback-Leibler divergence to accelerate convergence,  attentive experience replay (AER) selects experiences according to the similarities between their states and the agent’s current state \cite{SunZL20} and CER sets up two agents for competitive exploration between a pair of agents \cite{LiuTSX19}. But these three methods may rely on high computing resources. Hence, it is desirable to design a more effective and general approach to enhance the experience replaying method for DRL.
At the same time, quantum physics have been employed to dramatically enhance information processing capability \cite{shor1994algorithms,grover1997,Biamonte2017,Cai2015,Li2015,Beer2020,Bai2020,Ding2019} and have a positive influence on specific algorithmic tasks of applied artificial intelligence \cite{kak1995quantum,dunjko2018machine,Carleo602,Lloyd2014}.
In particular, there has been much interest in quantum enhancement of RL and their applications. The idea of quantum reinforcement learning was first originated from introducing the characteristics of quantum parallelism into classical RL algorithms \cite{Dong2008},
which achieved a better trade-off between exploration and exploitation
and sped up the learning as well. Quantum mechanics were found to be
able to bring an overall quadratic speedup for intelligent agents \cite{paparo2014quantum}.
The general agent-environment framework was also extended to the quantum domain \cite{Dunjko2016}. In addition,
quantum reinforcement learning with multi-qubits was evaluated on superconducting circuits \cite{Lamata2017} and was extended to other cases such as multi-level and open quantum systems \cite{cardenas2018multiqubit}.
Multiple value functions using Grover algorithm were proved to
converge in fewer iterations than their classical counterparts \cite{hu2019training}.
Recent research also demonstrated the advantage of RL using quantum Boltzmann machines over the classical one \cite{crawford2016reinforcement}.

%All the results suggest that the introduction of quantum characteristics into reinforcement learning is helpful and exhibit enhanced performance over various problems.

Inspired by quantum machine learning, we may produce atypical patterns in data.
For example, the quantum superposition state provides an exponential scale of computation
space in the $n$-qubits linear physical space \cite{Dong2008}, \cite{nielsen2010quantum}.
In this paper, we propose a quantum-inspired experience replay approach for deep reinforcement
learning (DRL-QER) to improve the training performance of DRL in a  natural way without
deliberate hyper-parameter tuning. In DRL-QER, the experiences are expressed in quantum
representations and the probability amplitudes of the quantum representations of experiences are iteratively manipulated by quantum operations, including the preparation operation and the depreciation operation. In particular, the preparation operation is designed according to the importance of the experiences and the depreciation operation is associated with the replaying times for the selected experiences. With the two operations, the importance of the experiences is  distinguished and the diversity of experiences is guaranteed. To test the proposed DRL-QER algorithm, experiments are carried out on
Gym-Atari platform with comparison to DRL-PER and DCRL. In addition, DRL-QER is implemented with double DQN and dueling DQN, and the DRL-QER variants are compared with their classical counterparts.

The rest of this paper is organized as follows. Section \uppercase\expandafter{\romannumeral2}
introduces DRL, experience replay, and the basic concepts of quantum computation as well.
In Section \uppercase\expandafter{\romannumeral3}, the framework of DRL-QER is introduced, quantum representations and quantum operations are presented, followed by the algorithm description of DRL-QER with specific implementation details. In Section \uppercase\expandafter{\romannumeral4},
experimental results are demonstrated to verify the performance of the proposed DRL-QER algorithm. Conclusions are drawn in Section \uppercase\expandafter{\romannumeral5}.

\section{Preliminaries}\label{Sec2}

\subsection{Deep Reinforcement Learning and Experience Replay}\label{Sec2.1}
\subsubsection{Markov Decision Process}
The training process of reinforcement learning (RL) is based on the model
of \emph{Markov Decision Process}, whose basic components can be described by a tuple
of $\langle S,A,P,R \rangle$ \cite{sutton2018reinforcement}, where $S$ is the
state space, $A$ is the action space, $P: S\times A\times S \to
[0,1]$ is the state transition probability and $R: S\times A\to
\mathbb{R}$ is the reward function.

In the process of interaction with the environment, the agent forms the state $s_{t}\in S$ at the
time step $t\in[0,T]$ and chooses an action $a_{t} = \pi(s_{t})$, $a_{t}\in A$ , where $T$ is the terminal time and policy $\pi$ is a mapping
from the state space $S$ to the action space $A$.  After carrying out the action
$a_{t}$, the agent transits to the next state $s_{t+1}$ and receives a scalar reward signal $r_{t}$. Thus we obtain a
transition of $e_{t}=(s_{t},a_{t},r_{t},s_{t+1})$ at the time
step $t$. RL aims at determining an
optimal policy $\pi^{*}$ so as to maximize the
cumulative discounted future rewards $R_t =
\sum_{k=0}^{T-t}\gamma^{k}r_{t+k}$, where $\gamma\in[0,1]$ is a
discount factor to balance the importance of the current rewards and
the future rewards. As a widely used RL algorithm, Q-learning defines $Q(s,a)$ as the expected discounted reward for
executing action $a$ at state $s$ following the policy $\pi$, and a look-up Q table is established to store the Q-values \cite{baird1995residual}.

%To solve reinforcement learning problems, the optimal state-action
%value function $Q^{*}(s,a)$ can be defined as the maximum
%expectation of return $R_{t}$ as follows:
%\begin{equation}\label{Eq1}
%    Q^{*}(s,a) = \max_{\pi} \mathbb{E}\{R_t|s_t=s,a_t=a\}.
%\end{equation}
%According to \emph{Bellman Equation}, \eqref{Eq1} can be rewritten as
%\begin{equation}\label{Eq2}
%    Q^{*}(s,a) = \mathbb{E}\{r+\gamma\max\limits_{a'}Q^{*}(s',a')|s,a\},
%\end{equation}
%where $s'$ is the next consecutive state and $a'$ is a legal
%action at $s'$. Q-learning \cite{watkins1992} is one of the widely
%used temporal difference methods to estimate $Q^{*}(s,a)$, whose
%one step updating rule is as follows:
%\begin{equation}\label{Eq3}
%   Q_{t+1}(s,a) = Q_{t}(s,a) + \alpha [r + \gamma \max_{a'} Q_{t}(s',a') - Q_{t}(s,a)],
%\end{equation}
%where $Q_{t}(s,a)$ denotes the state-action value function at time
%step $t$ and $\alpha\in[0,1]$ is learning rate. Under certain
%conditions \cite{sutton1998}, the convergence of Q-learning can be
%guaranteed, i.e., $t\to\infty$, $Q_{t}(s,a) \to Q^{*}(s,a)$.
%
\subsubsection{Deep Q Network}
In high-dimensional environments, it is a general and effective approach to approximate $Q(s,a)$ using a neural network with parameter $\theta$, i.e., $Q(s,a;\theta) \approx Q(s,a)$, instead of a look-up table that stores all state-action values $Q(s,a)$ \cite{baird1995residual}. In order to update the parameters of the neural network with a gradient descent method, the
``true values"  $y(s,a)$ of the
state-action values $Q(s,a)$ are estimated from the
maximum of the next state-action values $Q(s',a')$, i.e., $y(s,a) = r + \gamma\max\limits_{a'}Q(s',a';\theta^{-})$, where $\theta^{-}$ denotes the parameters of the target network
that are fixed during the computation of $y(s,a)$ and are updated after
some training steps.

The temporal-difference errors (TD-errors) $\delta$ can be measured by the deviation between $y(s,a)$ and $Q(s,a)$ as
\begin{equation}\label{Eq5}
 \delta = y(s,a) -  Q(s,a) = r + \gamma\max\limits_{a'}Q(s',a';\theta^{-}) - Q(s,a;\theta).
\end{equation}
Accordingly, the loss function $Loss(\theta; Q, y)$ to be optimized is
\begin{equation}\label{Eq6}
Loss(\theta; Q, y) = \frac{1}{2}(r + \gamma\max\limits_{a'}Q(s',a';\theta^{-}) - Q(s,a;\theta))^{2}.
\end{equation}
Differentiate the loss function $Loss(\theta; Q, y)$ with respect to the
parameter $\theta$, and we obtain the gradient as
\begin{equation}\label{Eq7}
\resizebox{.9\hsize}{!}{$\nabla_{\theta}Loss =[r + \gamma\max\limits_{a'}Q(s',a';\theta^{-})-Q(s,a;\theta)]\nabla_{\theta}Q(s,a;\theta)$}.
\end{equation}

\subsubsection{Experience Replay}
In most RL frameworks, agents incrementally update their parameters while they observe a stream of experiences. In the simplest form, the incoming data are used for a single update and discarded immediately, which brings two disadvantages: (\romannumeral1) strongly correlated transitions break the i.i.d. assumption which is necessary for many popular stochastic gradient-based algorithms; (\romannumeral2) the rapid forgetting of possibly rare experiences that are potentially useful in the future leads to sampling inefficiency. A natural solution would be to put the past experiences into a large buffer and select a batch of samples from them for training \cite{lin1992self}, \cite{Luo2018,Ni2019}. Such a process is called \emph{experience replay}.

In experience replay, how to choose the experiences (transitions) to be replayed plays a vital role to improve the training performance of DRL. When putting the transition $e_t$ into a fixed experience replay buffer with size $M$, a new index label, $k\in\{1,...,M\}$ is assigned to it, with its priority denoted as $P_k$. As such, the whole experience buffer can be regarded as a collection of transitions as $\{\langle e_t,P_k \rangle\}$. A complete process of experience replay is actually a store-and-sample process, and the learning process works by selecting a mini-batch samples from the whole buffer
to update the parameters of the RL agent. The key of experience replay lies in
 the criterion by which the importance of each transition is measured, i.e., to determine $P_k$ for each transition.

\subsection{Quantum Computation}\label{Sec2.3}
In quantum computation, the basic unit that carries information is a quantum bit (also called qubit) and a qubit can be in a superposition state of its eigenstates $|0\rangle$ and $|1\rangle$ \cite{nielsen2010quantum}, \cite{dong2010quantum},  which can be written as the following form of
\begin{equation}
|\psi\rangle=\alpha|0\rangle+\beta|1\rangle,
\end{equation}
where $\alpha$ and $\beta$ are complex numbers satisfying $|\alpha|^2+|\beta|^2 = 1$. Quantum mechanics reveals that measuring a qubit in the superposition state $|\psi\rangle$ leads it to collapse into one of its eigenstates of $|0\rangle$ with probability $|\alpha|^2$, or $|1\rangle$ with probability $|\beta|^2$. In particular, the coefficients can be written as $\alpha=\langle 0|\psi\rangle$ and $\beta=\langle 1|\psi\rangle$, where $\langle a | b\rangle$ represents the inner product between $|a\rangle$ and $|b\rangle$.

%Generally, any two-state quantum system could be regarded as a qubit, such as photons, atoms, electron spins, as well as nuclear spins. The main difference of state in quantum mechanics and classical ones is that the state of a classical system usually describes some real physical properties, which are generally observable, such as position and momentum, while it is impossible to directly observe a quantum state, since it has no correspondence to physical quantities of the quantum system. Also, the global phase of a quantum state has no observable physical effect, therefore $|\psi\rangle$ and $e^{i\alpha}|\psi\rangle$ (where $i=\sqrt{-1}$ and $\alpha \in \mathbb{R}$) could be regarded as the same quantum state ($|\psi\rangle \sim e^{i\alpha}|\psi\rangle$).

%\begin{figure}[!hb]
%  \centerline{\includegraphics[width=0.6\linewidth]{qubit.png}}
%  \caption{Qubit with arbitary superposition state. $|0\rangle$ and $|1\rangle $ correspond to two basis states for a qubit system. The state of such a qubit system could be formulated as $|\psi\rangle=\alpha|0\rangle+\beta|1\rangle$, where $\alpha$ and $\beta$ are complex coefficients, which are also called as probability amplitudes. Since the sum of probabilities must be equal to 1, $\alpha$ and $\beta$ should satisify $|\alpha|^2+|\beta|^2 = 1$}
%  \label{fig:qubit}
%\end{figure}

%However, we cannot determine in advance whether it will collapse to state $|0\rangle$ or $|1\rangle$.

In quantum computing, unitary transformation is an essential operation on quantum systems and can transform an initial state $|{\psi\rangle}$ to another state $|{\psi^{\prime}}\rangle$:
 \begin{equation}
|{\psi^{\prime}}\rangle = U|{\psi\rangle},
\end{equation}
where $U$ satisfies $U^{\dagger}U=UU^{\dagger} \equiv I$.
For example, a Hadamard gate that transforms $|0\rangle$ to $(|0\rangle+|1\rangle) / \sqrt{2}$ and $|1\rangle $ to $(|0\rangle-|1\rangle) / \sqrt{2}$ can be formulated as
\begin{equation}
H=\frac{1}{\sqrt{2}}\left[\begin{array}{cc}{1} & {1} \\ {1} & {-1}\end{array}\right].
\end{equation}
Another significant quantum gate is the phase gate, which is an important element to carry out the Grover iteration \cite{grover1997} for reinforcing the amplitude  of the “target” item. More discussions about quantum operations and quantum gates can be found in \cite{nielsen2010quantum}.

Grover algorithm is one of the most important quantum algorithms. It has been widely used in the problem of large scale database searching and is able to locate items with the complexity of $O(\sqrt N)$ in unstructured database with high probabilities. Its core idea is to represent items as a quantum system and manipulate its state using a  unitary operator in an iterative way \cite{grover1997}. As one of the main operations in Grover algorithm, Grover iteration has been successfully applied to RL methods \cite{Dong2008}, where the action is represented in the superposition of its possible eigen actions. Then, unitary transformation is iteratively performed on the superposition states to change the probability amplitudes of the ``good" actions.

%Similarly, we represent the experiences of agents by quantum states and attempt to evolve them by unitary operators during the whole learning process.

The state space of a composite quantum system is represented by the tensor product, denoted as $\bigotimes$, of the state space of each component system.
For example, the composite quantum system of two subsystems $A$ and $B$ can be defined on a Hilbert space
$\mathcal{H}=\mathcal{H}_{A} \otimes \mathcal{H}_{B}$, where $\mathcal{H}_{A}$ and $\mathcal{H}_{B}$ correspond to the Hilbert space of the subsystems $A$ and $B$, respectively.
Furthermore, its state $|\psi_{AB}\rangle$ may be described by the tensor product of the states of its subsystems, i.e.,
$|\psi_{AB}\rangle = |\psi\rangle_{A} \otimes|\psi\rangle_{B}$.
%For the composite quantum systems $A$ and $B$,

To obtain information by measuring or observing a quantum system, POVM (positive-operator-valued measure) can be applied \cite{nielsen2010quantum}. For an observable $H$, there exists a complete set of orthogonal projectors ${\left\{X_{h} : \sum_{h} X_{h}=I, X_{h}=X_{h}^{\dagger}, X_{\tilde{h}} X_{h}=\delta_{\tilde{h}, h} X_{h}\right\}}$, where $\delta_{\tilde{i}, i}$ is the Kronecker delta, $X_h$ is the projector onto the eigenspace of $H$ with eigenvalue $h$, and we have $H=\sum_{h} h X_{h}$. The probability of obtaining the outcome $h$ can be calculated by $P(h)=\left\langle\psi\left|X_{h}\right| \psi\right\rangle$.

\section{Deep Reinforcement Learning with Quantum-inspired Experience Replay}\label{Sec3}
In this section, the framework of DRL-QER is first introduced. Then quantum representations and quantum operations using Grover iteration are designed to provide a natural and appropriate experience replay mechanism. Finally, the implementation of the integrated DRL-QER algorithm is presented.
%To integrate quantum computation with deep reinforcement learning,
%the agent learns from past experiences  On one hand, the memory pool is represented as a composite system with each qubit in superposition state, then unitary transformation also called Grover iteration transformation is performed upon each qubit system for preparing states as well as updating states. Also, we perform select appropriate samples from experience pool based on quantum measurement principle. Finally, the implementation of the integrated DRL-QER algorithm is presented.

\subsection{Framework of DRL-QER}\label{Sec3.1}

In DRL-QER, quantum characteristics are borrowed to design new
manipulation methods to improve the experience replay mechanism,
which aims at providing a natural and easy-to-use experience replay approach using quantum representations and unitary transformation
for the experiences and their importance, respectively.

\begin{figure}[!hb]
  \centerline{\includegraphics[width=0.96\linewidth]{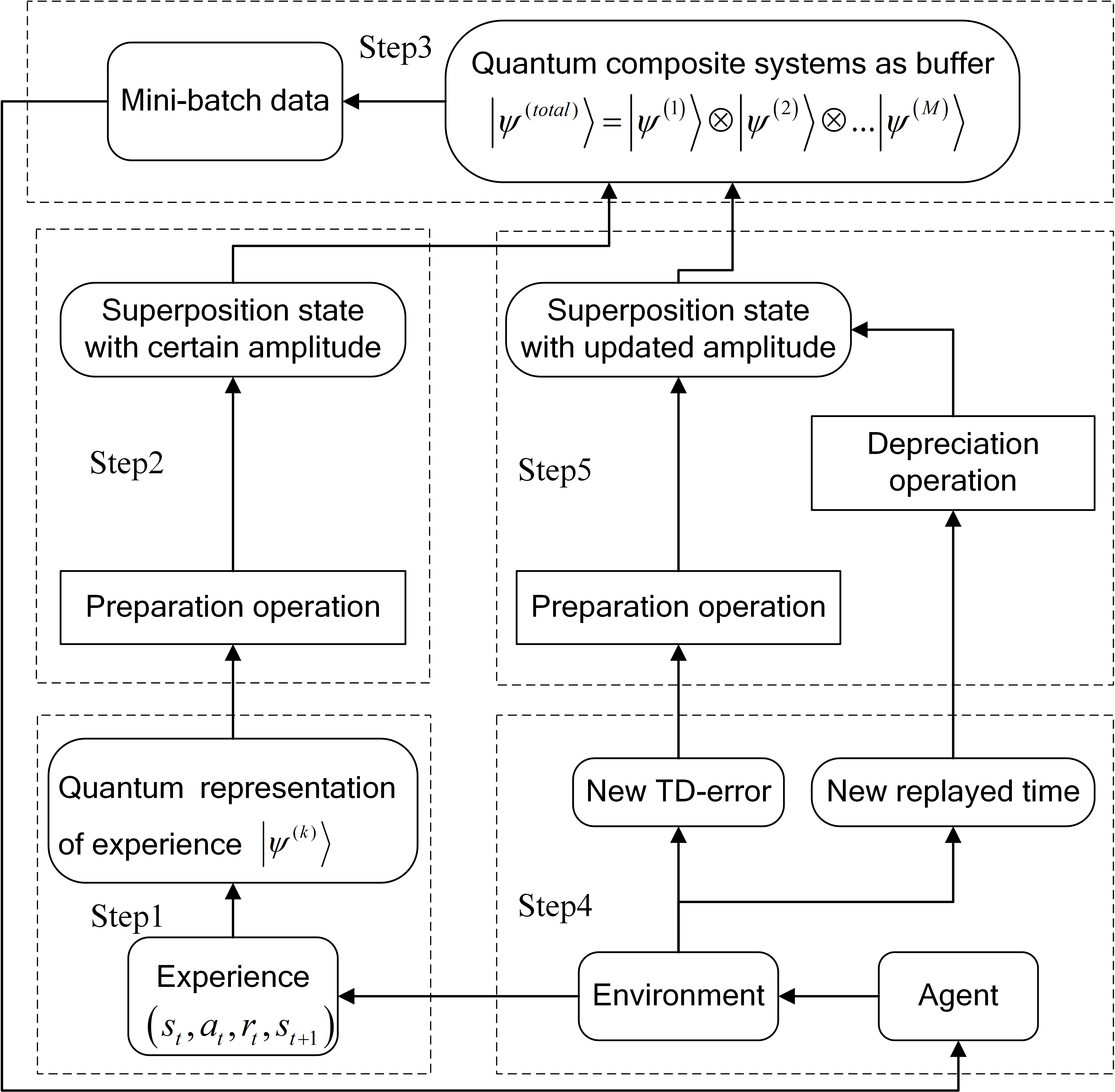}}
  \caption{Framework of DRL-QER. Step 1: representing the newly generated experience using a qubit; Step 2: performing the preparation operation on the quantum representation with Grover iteration; Step 3: sampling experiences to compose mini-batch data; Step 4: training the DRL agent with the mini-batch data; Step 5: updating the quantum representation of the experience by the new TD-error (the preparation operation) and the new number of replayed times (the depreciation operation) and then put it back into the replay buffer.}
  \label{fig:framework}
\end{figure}

The framework of DRL-QER is described as in Fig. \ref{fig:framework}. During each learning iteration, the agent
interacts with the environment and obtains the transition $e_t$ at time step $t$. Such a transition is first expressed in quantum representation, or more precisely, the $k$-th qubit, where $k$ is its index in the buffer. Secondly, the state of the qubit evolves to a superposition state through the preparation operation. Then, transitions are sampled with probabilities proportional to their importance and those selected samples compose the mini-batch data for training the neural network. In addition, after each training step, the amplitudes of the selected quantum representations are manipulated by the combined unitary transformation, including the preparation operation to adapt to the new TD-errors and the depreciation operation to deal with the replaying times of the transitions.
This procedure is carried out iteratively until the algorithm converges, whose specific details are implemented in the following subsections.

\subsection{Quantum Representation of Experiences}\label{Sec3.2}

In quantum theory, a qubit can be realized by a two-level atom, a spin system or a photon.
%the basic unit that carries information is a qubit, which can be in any superposition of its eigenstates, i.e., $|0\rangle$ and $|1\rangle$. The eigenstates for vector spaces usually have physical meanings for practical quantum systems.
For two-level atoms, $|0\rangle$ can be the ground state, while $|1\rangle$ represents the excited state.
For spin systems, $|0\rangle$ can be the state of \emph{spin up}, while $|1\rangle$ represents the state of \emph{spin down}. For photons, $|0\rangle$ can be the state of horizontal polarization, while $|1\rangle$
represents the state of vertical polarization. Here, in experience replay, one experience can be regarded as a qubit system, and its two eigenstates $|0\rangle$ and $|1\rangle$ represent the actions of \emph{rejecting} and \emph{accepting} this experience, respectively.

During the learning process, the agent tries to interact with its
environment, which can be modeled as an MDP. For each step $t$, with the current state $s_{t}$, the agent selects an action  $a_t$ under a certain exploration policy (such as $\epsilon$-greedy),
and then transfers to the next state $s_{t+1}$, and obtains a reward $r_{t}$.
Finally, four elements together compose a transition $(s_{t},a_{t},r_{t},s_{t+1})$, which is assigned a new index $k$ to denote its order in the experience buffer. In transforming the transition into quantum representation, we define the action of \emph{accepting} and \emph{rejecting} the transition as two eigenstates. Then, the transition is regarded as a qubit (as shown in Fig. 2) with its state as
\begin{equation}
|\psi^{(k)}\rangle=b_0^{(k)}|0\rangle+b_1^{(k)}|1\rangle,
\label{eq:state}
\end{equation}
where the coefficients $b_0^{(k)}$ and $b_1^{(k)}$ have probability amplitude meanings, and satisfy $|b_0^{(k)}|^2+|b_1^{(k)}|^2 = 1$. In particular, the probability of rejecting this transition is $|b_0^{(k)}|^2=|\langle 0|\psi^{(k)}\rangle|^2$ and the probability of accepting it is $|b_1^{(k)}|^2=|\langle 1|\psi^{(k)}\rangle|^2$. It is worthy to note that, the coefficients of the qubit are related with the significance of the experience. Before determining its importance, it is practical to first set an initial state, and let the qubit evolve from the initial state to a desired state.

In quantum computing, a uniform state is one significant superposition state, and has the form as
\begin{equation}
|\psi_{0}\rangle=\frac{\sqrt2}{2}(|0\rangle+|1\rangle).
\end{equation}
It has equal probabilities for two eigenstates and means that least knowledge is given about the state with a maximum entropy, which makes it feasible to adopt the uniform state as the initial state for each experience.

%which are usually given by TD errors and their values will be detailed in the following %subsection. In order to prepare each qubit system into a target state $|\psi^{(k)}\rangle=b_0^{(k)}|0\rangle+b_1^{(k)}|1\rangle$
\begin{figure}[!hb]
  \centerline{\includegraphics[width=0.6\linewidth]{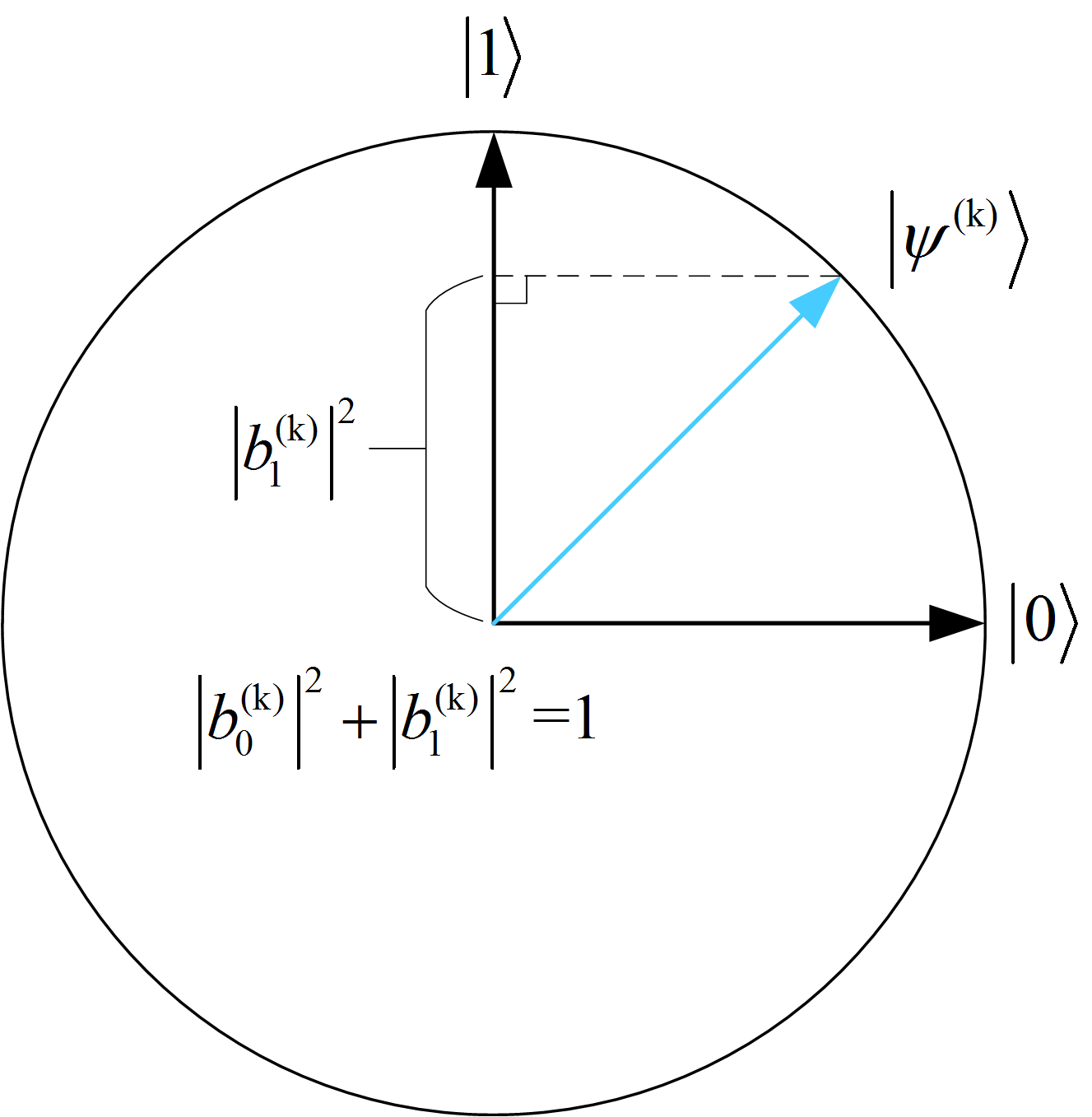}}
  \caption{An experience represented in a qubit system. Here, $|0\rangle$ and $|1\rangle $ correspond to \emph{rejecting} or \emph{accepting} the transition.
   The state of such a transition can be formulated as $|\psi\rangle=b_0^{(k)}|0\rangle+b_1^{(k)}|1\rangle$, where $|b_1^{(k)}|^2$ is the probability of accepting and $|b_0^{(k)}|^2$ corresponds to the probability of rejecting.}
  \label{fig:qubit}
\end{figure}

\begin{figure}[!hb]
  \centerline{\includegraphics[width=0.9\linewidth]{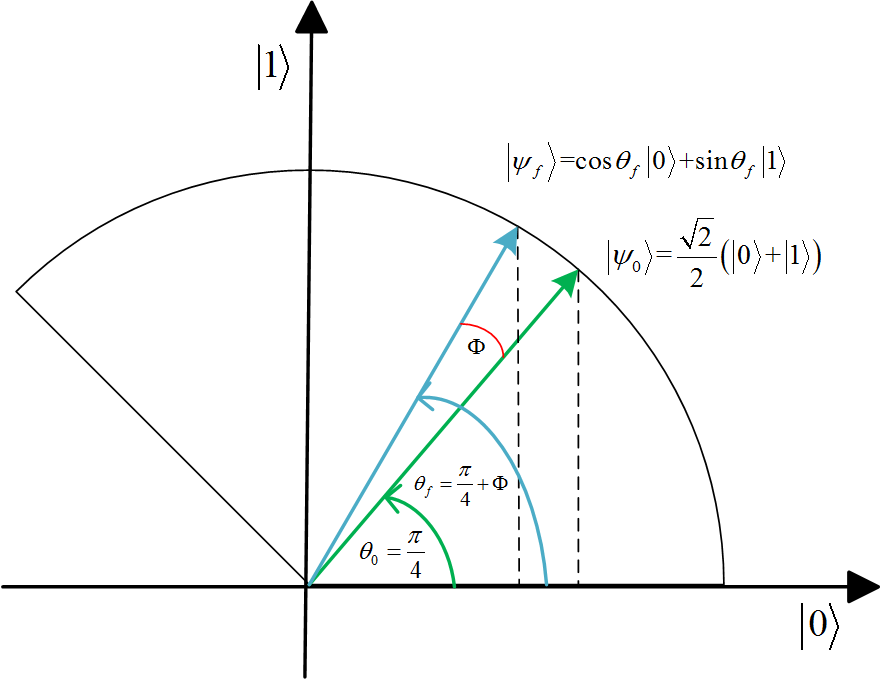}}
  \caption{ State transition of a qubit system under unitary rotation $U_{\Phi}$, where $|\psi_{0}\rangle$ is the initial state, $|\psi_{f}\rangle$ is the final state.
}
  \label{fig:Base transition}
\end{figure}

To adjust the probability amplitudes of the qubit state in (\ref{eq:state}), a rotation operator, which is the basic element of Grover iteration \cite{grover1997}, \cite{Dong2008}, is applied with
\begin{equation}
U_{\Phi}=e^{-\rm{i} \Phi Y}=
\left[
 \begin{matrix}
   \textup{cos}(\Phi) & -\textup{sin}(\Phi) \\
   \textup{sin}(\Phi) & \textup{cos}(\Phi) \\
  \end{matrix}
  \right],
  \label{eq:Grover}
\end{equation}
where $\Phi$ is a real number and has the physical meaning of rotation angle. Pauli $Y$ operator is given as
\begin{equation}
Y \equiv\left[\begin{array}{cc}{0} & {-\rm{i}} \\ {\rm{i}} & {0}\end{array}\right].
\end{equation}
The operation of performing a rotation operator $U_{\Phi}$ on a qubit of experience is visualized in Fig. \ref{fig:Base transition}. The quantum system evolves from the initial state $|\psi_0\rangle$ (the green one) to the final state $|\psi_f\rangle$ (the blue one), under the unitary transformation $U_{\Phi}$. Projecting $|\psi_0\rangle$ and $|\psi_f\rangle $ to the $y-$axis, the amplitude of observing $|1\rangle$ increases, which reveals that the probability of accepting the transition is slightly increased.

\noindent

Since the $k$-th experience in the buffer has the quantum representation form of  $\left|\psi^{(k)}\right\rangle $, the state of a memory buffer which is composed of $M$ experiences, is the tensor product of $M$ subsystems:
\begin{equation}
\left|\psi^{\text { total }}\right\rangle=\left|\psi^{(1)}\right\rangle \otimes\left|\psi^{(2)}\right\rangle \otimes \ldots\left|\psi^{(\mathrm{M})}\right\rangle.
\end{equation}

\subsection{Quantum Operations on Experiences}\label{Sec3.3}
%Similarly,  we  represent  the  experiences  of  agents  by quantum states and attempt to evolve them by unitary transformation during the whole learning process.
To deal with the quantum representations of experiences, three sub-processes are involved, i.e., preparation operation, depreciation operation and experience selection by quantum observation.
Firstly, the preparation operation is introduced to steer the quantum systems towards the target states, whose amplitudes are related to the TD-errors of the experiences. In fact, whenever the TD-errors of the experiences have changed, the preparation operation is performed to update their probability amplitudes. From this respect, every time when a suitable priority is determined, the quantum systems are to be transferred to a new target state, which can be regarded as a process of quantum state preparation. Hence, we call this special operation as the preparation operation. In addition, the depreciation operation is utilized to make sure that the significance of the experiences is adapted to the experience relaying process, such as the times of the experiences' being visited. Another significant operation is to select experiences by quantum observation, to compose a mini-batch data for training.

%For example, at the beginning, we could start from an initial state, e.g., the uniform state and then iteratively perform a unitary rotation on the systems until it evolves toward the target state where probability amplitudes are closely related with TD errors.
%To enhance the training efficiency,  Hence, once a mini-batch of experiences' quantum representations are selected and are fed into the neural network to train the agent.
To adjust the amplitudes of quantum systems in a natural and appropriate way, a Grover iteration method is adopted for both the preparation operation and the depreciation operation. Grover iteration is a significant operation for dealing with quantum states originated from classical information, and it aims at intensifying the probabilities of the target eigenstates, with others at equal probabilities \cite{grover1997}, \cite{Dong2008}. Considering that the probabilities of experiences' being extracted from an experience buffer vary, we do not use the conventional method, i.e., performing the unitary transformation on the composite system (the whole experience buffer). Instead, Grover iteration is conducted on a single experience with its quantum representation. This strategy helps to adaptively adjust the probability amplitude of each transition, and therefore to circumvent the neglect of the differences between experiences.

\subsubsection{Preparation operation}
To better optimize the process of experience replay in DRL-QER, the importance
of experiences needs to be distinguished first. Since a single rotation changes the probability amplitude of a qubit system, we define a basic rotation operator as

\begin{equation}
U_{\sigma}=e^{-\rm{i} \sigma Y}=\left[\begin{array}{cc}{\cos (\sigma)} & {-\sin (\sigma)} \\ {\sin (\sigma)} & {\cos (\sigma)}\end{array}\right],
\end{equation}
where $\sigma \in R$ is a tiny rotation angle. Based on the exponential approximation formula, i.e., $U_{\Sigma}=(U_{\sigma})^{m}$ with an integer $m$, several iterations of unitary rotations amount to an overall rotation on the qubits. In addition, owing to $e^{-\rm{i}\sigma Y} e^{-\rm{i}(-\sigma) Y}=I$, the rotation in the reverse direction can be conducted with $U_{\sigma}^{-1}$ (or $U_{\sigma}^{\dagger}$). Hence, different rotations can be achieved by performing multiple times of basic rotations in clockwise or counterclockwise directions.

%\begin{remark}
%Grover iteration method is widely used in the application of large scale database searching and is able to locate items with the complexity of $O(\sqrt N)$ in unstructured database with high probabilities. Its core idea is to represent items as quantum states, and then steer the quantum system by a diffusion unitary operator in an iterative way \cite{grover1996fast}. Grover iterations has been applied in reinforcement learning in \cite{Dong2008} due to its quantum properties and advantages, where the action is transformed to the superposition of $2^n$ possible eigen
%actions, and Grover iteration is performed on the superposition to change the probability amplitude of the action quantum system. Similarly, we represent the experiences of agents by quantum states and attempt to evolve them by unitary operators during the whole learning process.
%\end{remark}

%In our experiment, $M$ samples are dealt using the
%same technique since the experience buffer is usually composed of fixed samples. Hence,
%the whole buffer could be formulated as $M$ qubits systems with the probability or frequency
%of each transition determined by the principle of quantum measurement.
%we perform the measuring process and select appropriate samples from the experience pool. On basis of the  of each qubit system, we obtain the
%the more times a transition is replayed, the lower its value.

\begin{figure}[!hb]
  \centerline{\includegraphics[width=0.9\linewidth]{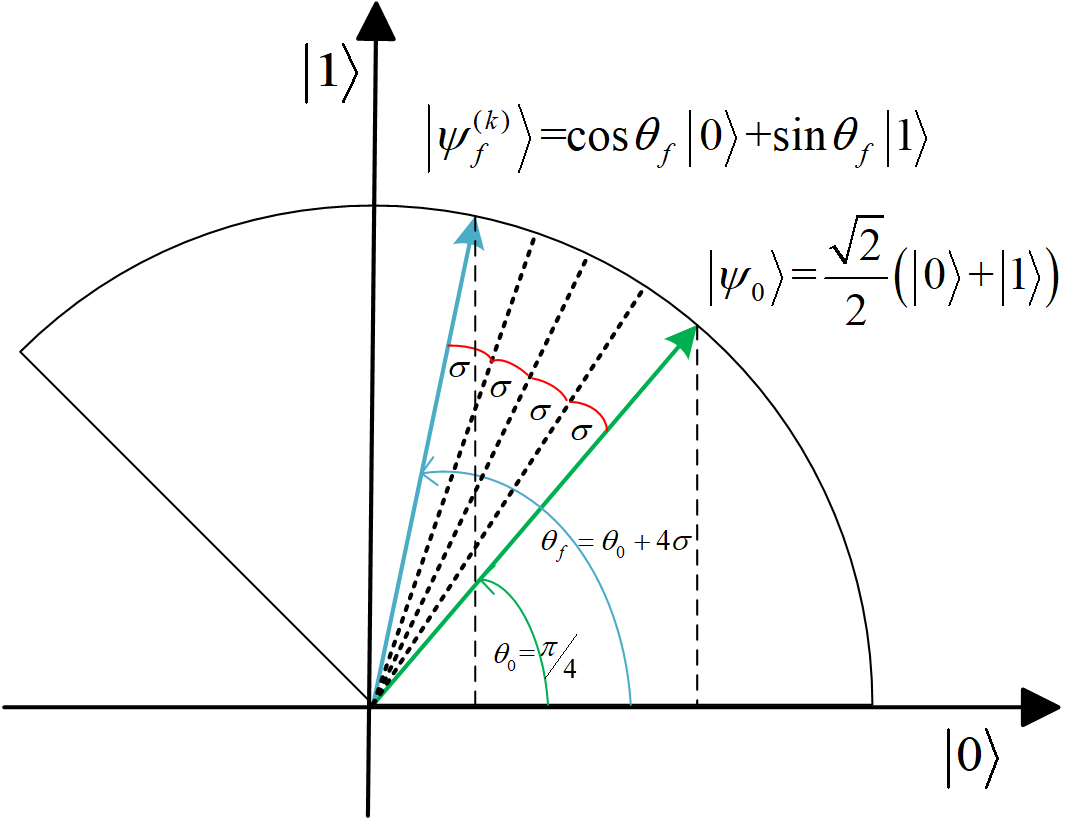}}
  \caption{The procedure of the preparation process using Grover iterations for the $k$-th qubit.  $|\psi_{0}\rangle$ is the uniform quantum state, $|\psi_{f}^{(k)}\rangle$ is the quantum state after conducting the transformation $U_{\Sigma}^{(k)}$.}
  \label{fig:grover iteration}
\end{figure}

%To achieve this, a basis rotation operator is firstly defined, and several times of basis rotation is performed on each transition, until it evolves towards a certain quantum state.To achieve this
The preparation operation for a single experience (the $k$-th transition in the buffer) is described in Fig. \ref{fig:grover iteration}, where four times of basic rotations in the counterclockwise direction are performed on the qubit to intensify the ``accepting" amplitude of the good experience or
equivalently to strength the ``rejecting" amplitude of the bad experience. Generally, the state evolution of such a quantum system can be expressed as
%\begin{equation}
%\begin{split}
%&U_{\Sigma}^{(k)}=\left\{\begin{array}{ll}{U_{\sigma}^{m_k}} & {\text { counterclockwise,}} \\ {U_{\sigma}^{-m_k}} & {\text { clockwise,}}\end{array}\right.\\
%& |\psi_{f}^{(k)}\rangle=U_{\Sigma}^{(k)} |\psi_{0}\rangle ,\\
%\label{eq:rotation}
%\end{split}
%\end{equation}
\begin{equation}
U_{\Sigma}^{(k)}=(U_{\sigma})^{m_k},\quad |\psi_{f}^{(k)}\rangle=U_{\Sigma}^{(k)} |\psi_{0}\rangle,
\label{eq:rotation}
\end{equation}
where $m_k$ represents the number of rotation times of the $k$-th qubit. Considering that TD-error reveals the importance of the transition, we associate the value of $m_k$ with its TD-error. For transition $e_t$ with TD-error $\delta_t$, the priority for $k$-th qubit is given as $P_k=|\delta_t|+\epsilon$ and the maximum priority of all the experiences is $P_{max}$. To convert the priority into the probability amplitude of the qubit, we try to map the priority $P_k$ to a rotation angle, which corresponds to a  unitary transformation of quantum states. In particular, $P_{max}$ is mapped to a rotation angle $\Sigma_{max}$, and then the angle of $P_k$ can be recorded as $\Sigma_k=\Sigma_{max} \times \frac{P_k}{P_{max}}$. Since Grover iterations aim at iteratively performing unitary transformations until a desired state is achieved, $\Sigma_{max}$ is split into $\mu$ pieces, and each piece is assigned with $\sigma$. The initial state of the qubit is assigned with rotation angle $\iota$, so the angle of rotation can be defined as the target angle minus the initial angle. Finally, the value of $m_k$ reads as
\begin{equation}
m_k=\textup{Floor}(\mu \times P_k/P_{max}-\iota /\sigma),
\label{eq:mk}
\end{equation}
where $\mu,\iota \in R$ are two hyper-parameters, and  $\textup{Floor}(x)$ takes the largest integer not greater than $x$. In particular, the sign of $m_k$ reflects the rotation direction relative to the angle of the uniform state, i.e., $\frac{\pi}{4}$. For example, when $m_k$ is a positive integer, Grover iteration works in the counterclockwise direction, otherwise, it is conducted in the clockwise direction.

The value of $\sigma$ in (\ref{eq:rotation}) is usually carefully set since it plays an important role in the quantum representation of experiences. From a convenient point of view, it is the most appropriate to set a fixed value. While from the perspective of adaptation to different environments, associating it with the training process, such as the TD-errors, the maximum times of experiences' being visited and the training steps is more preferable. In this work, we describe it with a function associated with the training episode $TE$:
\begin{equation}
\sigma= \frac{\zeta_1}{1+e^{\frac{TE}{\zeta_2}}},
\end{equation}
where $\zeta_1, \zeta_2 \in R$ are two hyper-parameters.
%The iteration times $m_k$ for $k-$th experience with TD -error $\delta_k$ is given as
%\begin{equation}
%m_k=\mu_1 \times \delta_k/\delta_{\textup{max}}-\mu_2 /\sigma,
%\end{equation}
%\noindent
%where $\delta_{\textup{max}}$ is the maximum value of TD-errors among the experience buffer.
% \begin{remark}
%  In (14), the right part of the equation can be clarified as Floor$((\mu_1\times\sigma \times P_k / \delta_{max}-\mu_2)/\sigma)$, where $P_k / \delta_{max}$ is the ratio of the priority of the specific experience to the maximum TD-error, $\mu_1\times \sigma$ represents the angle of the maximum TD-error mapping, then it minus the $\mu_2$ (which is the angle of uniform state) to compute the rotation angle. Considering  that the Grover iterations are a series of discrete operations, the rotation angle should be split into small pieces, so the $\sigma$ is divided. Here $\mu_1$ is related to the accuracy of Grover iteration, and $\sigma$ decides the angle of the maximum TD-error mapping to,  and further determine which experiences will be sampled with the greatest probabilities. 
% \end{remark}

By performing the same procedure to each transition, all experiences will end up in their target quantum representations. For example, for most of those valuable transitions, performing the preparation operation in the counterclockwise direction makes them approach $|1\rangle$, while for those less important experiences, the preparation operation in clockwise direction can be deployed on their quantum representations to make them closer to $|0\rangle$.

% Finally, all the experiences are attached with different importance.
%this operation is called the \emph{preparation} Grover iteration

% \begin{remark}
%  Here preparation operation is designed to optimize the priorities of the experiences. The experiences with both tiny and large TD-errors compose a dataset (or experience buffer) which is challenging or noisy. As is mentioned above, if the experiences are sampled in the same way as PER, some overused experiences would contribute large gradients in training, which would cause the neural network to oscillate. From this perspective, in preparation operation experiences with the largest sampling priorities are those with larger but not the largest TD-errors.
% \end{remark}

%The problem is that no matter how many times they are replayed, their TD-errors would never %reduce.
\subsubsection{Depreciation operation} After the process of preparation, the probabilities of selecting the experiences are closely associated with their TD-errors. However, in actual training, some experiences are replayed at high frequencies and may result in poor learning performance, which is called over-training, and the  limited size of the replay buffer may aggravate this situation \cite{de2015importance}. In RL, over-training reveals the issue of exploration–exploitation tradeoff
\cite{ishii2002control,abbeel2005exploration,mannucci2017safe,Dong2012robust,Chen2014Fidelity,Zhang2019}. Sufficient
exploration in the state-action space helps prevent
the algorithm being trapped in locally optimal solutions, while
exploiting the current policy helps the algorithm converge as
fast as possible. To achieve a balance between exploration and exploitation, the sample diversity is considered to enhance the learning performance of the agent. As such, the depreciation operation is developed for the experiences according to the replaying process. This is achieved by iteratively modifying their probabilities once the transitions are selected, whose effect contains and is greater than the utilization of importance-sampling correction, which is demonstrated in the ablation experiments in the supplementary material.

%For the sake of generalization of reinforcement
%learning agent, we take into account the sample diversity in  DRL-QER. %Accordingly, if the agent is restricted to
%some specific transitions, it cannot see the wood for the trees.
%By contrast, there is no need to manage all aspects of the
%environment.

Once the experiences are selected and put back in the memory buffer for training, their importance to the agent is unavoidably changed,
not only because their TD-errors have been changed, but also in that they are no longer
brand new to the agent. Therefore their probability  amplitudes need to be modified.
From this perspective, another unitary transformation 
\begin{equation}
U_{\omega}=\left[\begin{array}{cc}{\cos (\omega)} & {-\sin (\omega)} \\ {\sin (\omega)} & {\cos (\omega)}\end{array}\right]
\end{equation}
\noindent
is used for the depreciation operation, with $\omega \in R$. In particular, it is implemented on the selected experiences using Grover iteration. Every time the experiences have been accepted, their quantum representations go through a unitary transformation as follows \begin{equation}
|\psi^{(k)}_f\rangle \leftarrow U_{\omega}|\psi^{(k)}_f\rangle.
\end{equation}

The value of $U_{\omega}$, or more precisely $\omega$, should be adapted to the specific scenario.
In experience replay, when the buffer is full, new transitions are orderly put in the buffer,
with the old ones replaced. Besides, the period of the experiences' being replaced is a fixed
number of steps. Hence, a transition will be kept in the buffer for fixed time steps, before it is replaced.
In that case, during fixed training steps, the total replaying times of all the experiences are fixed.
A large value of the maximum number of replaying times among all the experiences (denoted as $RT_{\max}$)
reveals an uneven replaying distribution, which means that some experiences have outstanding
priorities compared to other experiences. 
To weaken this phenomena, a smaller $\omega$ helps to retain those less important experiences; otherwise, a large depreciation factor might result in sharp declines in the accepting probabilities of those experiences. Hence, the value of $\omega$ is decreased with $RT_{\max}$.

In addition, $\omega$ should be adapted to the training episode $TE$. In the early training stage, the importance of  experiences is ambiguous. While, after a period of training, the TD-errors of some experiences tend to remain in large values, no matter how many times they have been selected to update the network. Therefore, it is feasible to ``intensify" the accepting probabilities of the experiences that have been replayed with more times compared with others at the early training stage and to ``cool'' down their accepting probabilities to avoid over-training at the later stage. This is realized by increasing $\omega$ with the training episode $TE$.
Finally, the depreciation factor $\omega$ is given as
\begin{equation}
\omega=  \frac{\tau_1}{RT_{\max}(1+e^{\tau_2/TE} )},
\label{eq:Depreciation}
\end{equation}
where $\tau_1, \tau_2 \in R$ are two hyper-parameters.

% \begin{remark}
% Here the depreciation operation plays two roles: 1) prevent the experiences whose large TD-errors do not decrease after numbers of replays from overusing, guarantee the diversity of experiences. 2) Act as importance sampling correction \cite{schaul2015prioritized} to make learning less aggressive, especially for those gradients contributed by experiences with large TD-errors. Considering the second point, we do not deploy importance sampling correction in the DRL-QER to avoid repeated effects. 
% \end{remark}

%\begin{remark}
%Here the size of experience buffer should also be considered in (\ref{eq:Depreciation}). Its impact lies in that, the larger the experience buffer, the larger the period for updating experience, and the slower the "invalid experience" is discarded,  so the depreciation factor should be set larger  to balance this effect.
%\end{remark}
%Observation is first performed on all the experience in the experience buffer, then there will
%be an uncertain amount of experience being selected. The last progress is randomly selecting 32
%piece of experience from them, as is shown in Fig. \ref{fig:Observation process of experience replay}
%
%\begin{figure}[!hb]
%  \centerline{\includegraphics[width=0.7\linewidth]{observation_process_of_experience.png}}
%  \caption{Observation process of experience replay}
%  \label{fig:Observation process of experience replay}
%\end{figure}

\subsubsection{Experience selection by quantum observation} To accomplish the training process,
samples are chosen from the buffer and fed into the network for learning. Here, we draw from
the quantum measurement principle and determine the probabilities of experiences based on
quantum observation. For the $k$-th qubit in state $|\psi^{(k)}_f\rangle$,
observing its probability of being accepted is $|\langle 1| \psi^{(k)}_f\rangle|^2$, which
is actually the probability of measuring $|1\rangle$. Then, normalizing the probability
based on all transitions, we obtain its replaying probability as
%\begin{equation}
%b_k=\frac{|b_1^{(k)}|^2} {\sum_{i}|b_1^{(i)}|^2}.
%\end{equation}

\begin{equation}
b_k=\frac{|\langle 1| \psi^{(k)}_f\rangle|^2} {\sum_{i}|\langle 1| \psi^{(i)}_f\rangle|^2}.
\label{eq:frequency}
\end{equation}

%then normalized based on all transitions, to obtain their frequencies.

\begin{figure}[!hb]
	\centerline{\includegraphics[width=0.96\linewidth]{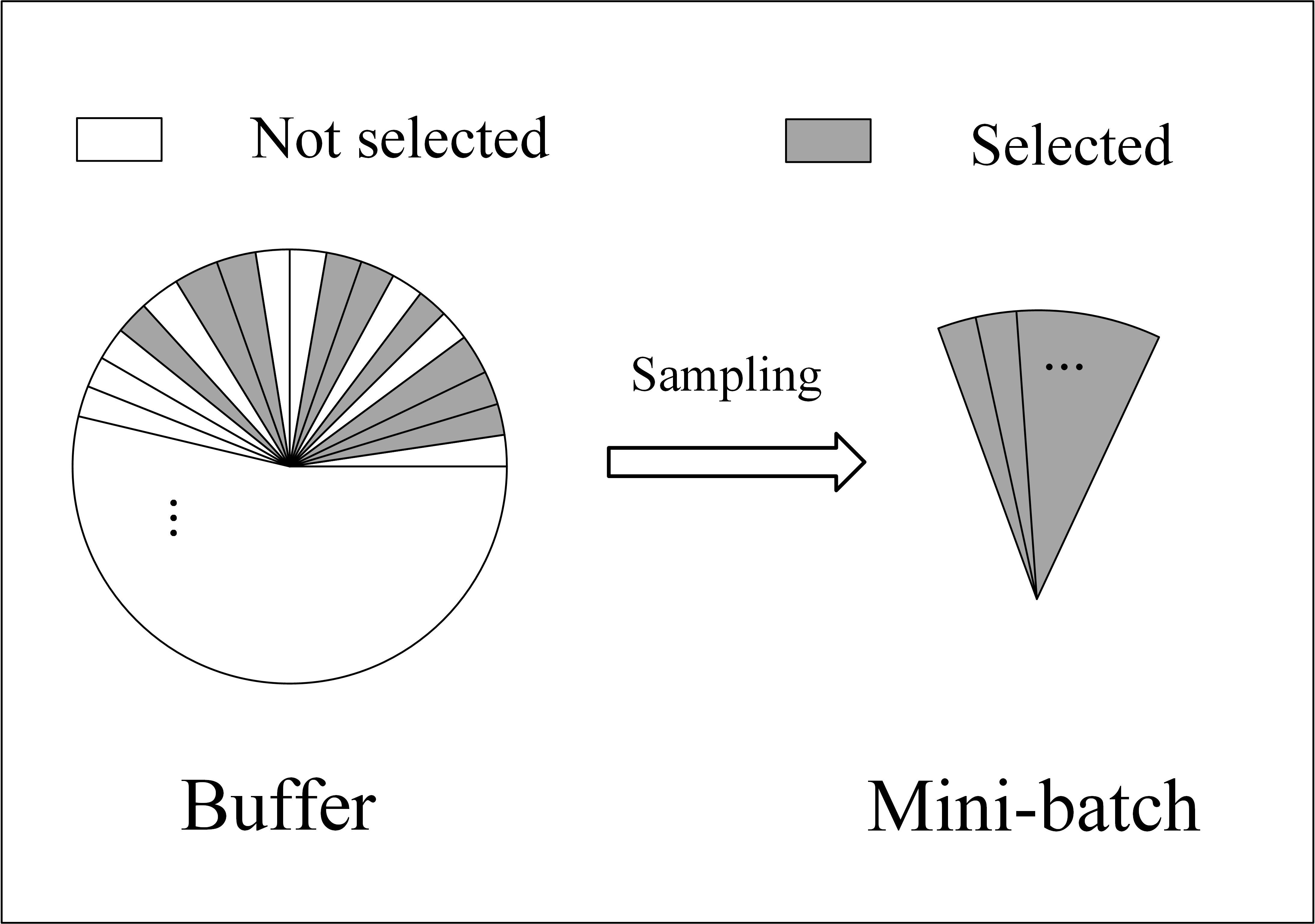}}
	\caption{The observation process for experience replay. The buffer is composed of a number of transitions, where each one is accompanied with its probability drawn from quantum observation principle. The transitions are sampled out from buffer according to their replaying probabilities to compose the mini-batch data.}
	\label{fig:Observation process of experience replay}
\end{figure}

The process of experience selection is summarized as in
Fig. \ref{fig:Observation process of experience replay},
where each transition has its own probability in the buffer.
Inspired by the quantum observation principle, this process determines the probabilities of being selected among the buffer. During the sampling process, several times of sampling one transition from the experience buffer are performed under fixed probabilities. The sampling times are consistent with the size of minibatch, which is set as $32$ in the simulations.

\begin{remark}
The process of obtaining minibatch data from the buffer with fixed probabilities is a sampling process with replacement. For each sampling process, the selected sample is still retained in the buffer and is reset to the uniform state after being sampled. This idea is inspired from the phenomena that observing a quantum system makes its state collapses. In that case, the quantum operation (i.e., the preparation operation and the depreciation operation) on the selected quantum experience starts from the uniform state, rather than its previous state.
\end{remark}

\subsection{Implementation}\label{Sec3.4}

\begin{algorithm*}
	\caption{DRL-QER Algorithm}\label{Algorithm2}
	\KwIn {size of experience buffer $M$, size of mini batch $N$.}
	Initialize the preparation factor $\sigma$,  the depreciation factor $\omega$, the maximum TD-error $\delta_{\max}$, the replayed time vector $cn = [cn_{1},cn_{2},\ldots, cn_{M}]=\vec{0}$, the index in the buffer $k=1$, a variable $LF=False$\;
	\For{$TE=1 \to Training Frames$}
	{
		Observe $s_{1}$ and choose $a_{1}\sim\pi(s_{1})$\;
		\For{$t=1 \to T$}
		{
			Observe $r_{t},s_{t+1}$ and then obtain a transition $e_t$\;
			\If{$s_{t+1}$ is terminal}
			{break;}
			
			Initialize the $k$-th qubit as the uniform state $|\psi_0\rangle$\;
			Set $P_k=|\delta_{max}|$ and obtain $m_k$ according to (\ref{eq:mk})\;
			Perform the preparation operation on $k$-th qubit using Grover iteration, and obtain its final state $|\psi^{(k)}_f\rangle=(U_{\sigma})^{m_k} |\psi_0\rangle$\;
			
			Store the transition $e_t$ with its quantum representation $|\psi^{(k)}_f\rangle$ in the buffer\;

			\If{LF==True}
			{Determine the probabilities of the experiences by quantum observations and obtain their replaying probabilities $\left[b_{1},b_{2},...,b_{M}\right]$ according to (\ref{eq:frequency})\;
			Update the preparation factor $\sigma$ and the depreciation factor $\omega$\;
						\For{$j=1\to N$}
				{
							Sampling a transition  with its index in the buffer as $d\in\{1,2,...,M\}$ based on $\left[b_{1},b_{2},...,b_{M}\right]$\;
							Reset the $d$-th qubit back to the uniform state $|\psi_0\rangle$\;
							Compute its TD-error $\delta_j = r_{j} + \eta\max_{a}Q_{target}(s_{j+1}, a) - Q(s_{j},a_{j})$\;
							Obtain its priority $P_d=|\delta_j|$ and obtain $m_d$ according to (\ref{eq:mk})\;
							Update the replaying time $cn_{d}$ by $cn_{d}= cn_{d} + 1$\;
							
							Conduct a complex Grover iteration process including both the preparation operation and the depreciation operation on the experience's quantum representation $|\psi^{(d)}_f\rangle = (U_{\omega})^{cn_d} (U_{\sigma})^{m_d} |\psi_0\rangle $\;
							Update $\delta_{max}=\max(\delta_{max},|\delta_{j}|)$ and update $RT_{\max}=\max(cn_1,cn_2,...,cn_M)$\;

				}	
				
			Update weights $\theta$ by stochastic gradient descent\;
			Copy weights into target network $\theta_{target} \leftarrow \theta$\;
			Remove the $k$-th quantum representation of experience from the buffer and reset $cn_k=0$\;	
						
			}
			
				$k\leftarrow k+1$\;
				\If{$k > M$}
				{ Set $LF=True$ and set $k=1$;}

			Choose action $a_{t+1} \sim \pi(s_{t+1})$\;
		}
	}
\end{algorithm*}

An integrated DRL-QER algorithm is shown as in Algorithm
\ref{Algorithm2}. During each step, the agent
encounters transition $e_{t}$. Considering that the newly generated transition
does not have a TD-error, we assign the maximum TD-error for it, i.e., $\delta_t=\delta_{\max}$
to give it a high priority. This guarantees that every new experience is sampled with a high priority.
Then, the transition is represented as a qubit, with its initial state as $|\psi_0\rangle$.
The preparation operation using Grover iteration is performed on the experience until it reaches the
final state $|\psi^{(k)}_f\rangle$. After the buffer is full, transitions are sampled
with probabilities proportional to their amplitudes of quantum states, and those selected
samples compose the mini-batch data for training the network. For the selected transitions,
after being reset to the uniform state, their corresponding quantum representations are manipulated
through the preparation operation to adapt to new priorities, and the depreciation operation to
adapt to the replaying times.
This procedure is carried out iteratively until the algorithm converges.

%Finally, replayed time factor $cn$, preparation factor $m$,
%TD-errors $\delta$ and quantum states $|\psi\rangle$ are updated according to the above principles.

%\begin{remark}
%The proposed algorithm is under the basic framework of DQN, and %teratively optimizes the process of experience replay by drawing from %Grover iteration method.
%The learning effect of DRL-QER is similar to that of DQN, PER and %other algorithms, so there will be no big oscillations during the %learning process.
%\end{remark}

%In our learning process, the textbook should be updated every few decades, similarly in the

\begin{remark}
 The proposed QRL-QER method works by representing the classical information (experiences) into quantum forms and performing quantum operations. Although it is inspired by quantum laws, the process can be simulated on a classical device. Hence, it is a quantum-inspired algorithm, and does not need to be implemented on a quantum device.
\end{remark}

\begin{remark}
In DRL, the buffer is to store the past experiences and use them to update the parameters
of the agent. During this process, the agent interacts with the environment under the new
network parameters. Hence, the experiences in the buffer should be updated after some
training steps to gain a better training effect. To achieve this, the buffer is set
as a fixed size and the oldest experience is discarded to make room for the newly
produced experience (reset $k=1$ in Algorithm \ref{Algorithm2}) when the buffer is
full ($k>=M$ in Algorithm \ref{Algorithm2}). In addition, the procedure of updating
the parameters of the network begins after the buffer is full, i.e., the variable $LF$ is set $True$. This technique is also applied to DRL-PER and DCRL to achieve a fair comparison in the following experiment section. 
In the implementation of Algorithm 1, we set a predefined value for the maximum value of TD-error, i.e., $\delta_{\max}$. During the whole learning process, $\delta_{\max}$ should be updated once a larger TD-error is found. As such, new $\delta_{\max}$ is assigned to the future newly generated transitions to give them the highest priorities.
\end{remark}

\section{Experiments}\label{Sec4}
To test the proposed DRL-QER algorithm, several groups of experiments are carried out
on Atari games with comparison to two benchmark algorithms (DRL-PER and DCRL). In addition, DRL-QER is combined with double network and dueling network, and tested on additional experiments to verify its performance.

%At the same time, some important indicators are also listed in the experimental results.

\subsection{Setup}\label{Sec4.1}
The experiments are carried out on the widely used platform OpenAI Gym to play Atari 2600 games
\cite{Gym2016},
and the testing games can be divided into four categories, namely shooting games,
 antagonistic games,  racing
games and  strategy games. For all the games, the agent takes high-dimensional data ($210\times 160$ colour video) as input to learn good policies. In order to win the games, the agent has to  plan over the long-term. All the experiments are deployed on a
computer of ThinkStation P920 with 24xCPU@2.40GHz, Nvidia Tesla
p5000, Ubuntu 16.04.5 LTS, Python.

% To verify the effectiveness of  DRL-QER, two baseline algorithms including DQN \cite{mnih2015human} and
% DRL-PER \cite{schaul2015prioritized} are also tested for comparison. The sampling method in  DRL-QER  can be regarded as a generalization of those in DQN and DRL-PER. DQN disregards the differences in the importance
% of experiences and treat them equally. It can be considered as a case of DRL-QER
% in which quantum representations remain in the uniform states without any transformation.
% DRL-PER samples  experiences according to their TD-errors, and this can be regarded as a situation in which DRL-QER
% does not consider the influence of over-training and discards the depreciation operation.

To verify the effectiveness of  DRL-QER, two baseline algorithms including 
DRL-PER \cite{schaul2015prioritized} and DCRL \cite{ren2018} are also tested for comparison. The sampling method in  DRL-QER  can be regarded as a generalization of that in DRL-PER. 
DRL-PER samples  experiences according to their TD-errors with proportional prioritization, and this can be regarded as a situation in which DRL-QER
does not consider the influence of over-training and discards the depreciation operation.

When deploying DRL-QER on Atari 2600 games, we adopt a similar neural network architecture and a same hyper-parameter setting to those in \cite{mnih2015human} and \cite{schaul2015prioritized}. Considering the hardware limitation, the high computational requirement and the high time cost, we make some fine-tuning on the hyper-parameters. In order to avoid the expensive cost of training for 50 million frames,
we train for 5 million frames.
Generally, it is hard and not necessary to accomplish quantum operations on a classical computer. We need to do necessary approximation in the simulation of DRL-QER, i.e.,
the states of qubit systems are represented by two-dimensional complex vectors, and the preparation operation and the depreciation operation are performed in the form of unitary matrix transformation.
 In addition, the normalized probabilities are stored in a special binary heap called ``sum tree", where the value of a parent node is the sum of all values of its children.
Last but not least, for performing necessary operations on experiences, we introduce some hyper-parameters for the preparation factor
$\sigma$ and the depreciation factor $\omega$, and their values are provided in Table \ref{Table1}.
Other hyper-parameters of DRL-QER are selected by performing a grid search on the game Breakout.
%\ref{Table2} only shows partial results of grid search.

%rotations in two-dimensional coordinate systems

\begin{table}[!htbp]
\renewcommand\arraystretch{2}
\centering \caption{Hyper-parameters Adjustments in Numerical
Experiments} \label{Table1}
\begin{tabular}{|c|c|c|c|}
\hline
\multicolumn{4}{|c|}{Altered Hyper-parameters}\\
\hline
\multicolumn{2}{|c|}{Hyper-parameter} & Original Value & Altered Value\\
\hline
\multicolumn{2}{|c|}{Training Frames} & $5\times 10^7$ & $5\times 10^6$\\
\hline
%\multicolumn{2}{|c|}{Importance Sample Variable $\beta$} & $0.4 \rightarrow 1.0$ & 0.4\\
%\hline
\multicolumn{2}{|c|}{Preparation sub-factor $\zeta_1$} & ---  & $0.03\pi$\\
\hline
\multicolumn{2}{|c|}{Preparation sub-factor $\zeta_2$} & ---  & $2\times10^6$\\
\hline
\multicolumn{2}{|c|}{Depreciation sub-factor $\tau_1$} & ---  & $\pi$\\
\hline
\multicolumn{2}{|c|}{Depreciation sub-factor $\tau_2$} & ---  & $1\times 10^6$\\
\hline
\multicolumn{2}{|c|}{Parameter $\mu$ of $m$} & ---  & 100\\
\hline
\multicolumn{2}{|c|}{Parameter $\iota$ of $m$} & ---  & $0.25\pi$\\
\hline

\end{tabular}
\end{table}

\begin{table}[!htbp]
\renewcommand\arraystretch{2}
\centering \caption{Average Rewards Per Episode of DRL-PER, DCRL and
DRL-QER} \label{Table2} \scalebox{0.95}{
\begin{tabular}{|p{2cm}<{\centering}*{3}{p{1.8cm}<{\centering}}|}
\hline
Game Name & DRL-PER($\pm$std) & DCRL($\pm$std) & DRL-QER($\pm$std)\\
\hline
Alien & 1270.2($\pm341.3$) & 1223.9($\pm297.1$) & \textbf{1309.3}($\pm348.8$)\\
\hline
Beam Rider & 1448.3($\pm290.2$) & \textbf{1594.6}($\pm302.3$) & 1508.2($\pm330.4$)\\
\hline
Breakout & 5.7($\pm2.1$) &5.2($\pm1.9$) &  \textbf{5.8}($\pm2.8$)\\
\hline
Carnival & 1142.5($\pm376.2$) & \textbf{1235.1}($\pm331.5$) & 1214.0($\pm413.4$)\\
\hline
Enduro & 43.0($\pm20.6$) & \textbf{44.3}($\pm20.5$) & 42.7($\pm19.2$) \\
\hline
Freeway & 60.6($\pm6.6$) & 60.2($\pm5.8$) & \textbf{60.7}($\pm6.3$)\\
\hline
Kangaroo &1129.3($\pm291.3$) & \textbf{1237.3}($\pm303.2$) & 1143.3($\pm324.6$)\\
\hline
Kung-Fu Master & 521.3($\pm328.8$) & 670.0($\pm434.8$)  & \textbf{712.0}($\pm370.9$)\\
\hline
Ms. Pacman & 1862.2($\pm661.8$) & 1918.9($\pm741.2$) & 1903.5($\pm735.3$)\\
\hline
River Raid & \textbf{2248.6}($\pm626.8$) & 1239.5($\pm251.6$) & 1479.5($\pm307.5$)\\
\hline
Road Runner & 2312.7($\pm897.7$) & \textbf{3615.3}($\pm1543.8$) & 3208.7($\pm1298.7$)\\
\hline
Space Invaders & 679.2($\pm321.2$) & 735.1($\pm250.8$) & \textbf{741.5}($\pm317.2$)\\
\hline
\end{tabular}}
\end{table}

\begin{table}[!htbp]
\renewcommand\arraystretch{2}
\centering \caption{Average Rewards Per Episode of DRL-PER with double network and
DRL-QER with double network} \label{Table3} \scalebox{0.95}{
\begin{tabular}{|p{2.4cm}<{\centering}*{2}{p{2.5cm}<{\centering}}|}
\hline
Game Name & DRL-PER($\pm$std)  & DRL-QER($\pm$std)\\
\hline
Alien & 1215.2($\pm392.0$) & \textbf{1275.0}($\pm350.0$)\\
\hline
Carnival & 2420.5($\pm315.2$) & \textbf{2480.3}($\pm319.2$)\\
\hline
River Raid & \textbf{1906.2}($\pm522.6$) & 1870.9($\pm310.7$)\\
\hline
Space Invaders & 723.6($\pm229.9$) & \textbf{750.3}($\pm262.8$)\\
\hline
\end{tabular}}
\end{table}

\begin{table}[!htbp]
\renewcommand\arraystretch{2}
\centering \caption{Average Rewards Per Episode of DRL-PER with dueling network and
DRL-QER with dueling network} \label{Table4} \scalebox{0.95}{
\begin{tabular}{|p{2.4cm}<{\centering}*{2}{p{2.5cm}<{\centering}}|}
\hline
Game Name & DRL-PER($\pm$std)  & DRL-QER($\pm$std)\\
\hline
Alien & 801.4($\pm131.6$) & \textbf{854.6}($\pm140.3$)\\
\hline
Carnival & 1317.0($\pm376.5$) & \textbf{1367.7}($\pm380.2$)\\
\hline
River Raid & \textbf{3249.2}($\pm522.0$) & 3007.3($\pm456.7$)\\
\hline
Space Invaders & 756.1($\pm229.5$) & \textbf{758.8}($\pm262.2$)\\
\hline
\end{tabular}}
\end{table}

\begin{figure*}[!htbp]
\begin{minipage}{0.3\linewidth}
  \centerline{\includegraphics[width=2.4in,height=2in]{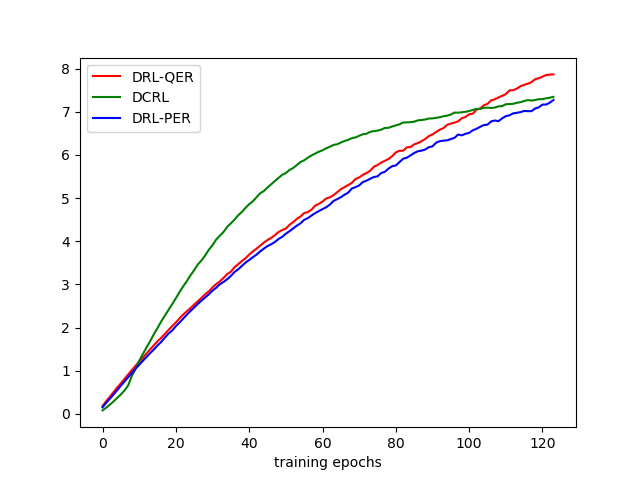}}
  \centerline{ (a) Space Invaders }
\end{minipage}
\hfill
\begin{minipage}{0.3\linewidth}
  \centerline{\includegraphics[width=2.4in,height=2in]{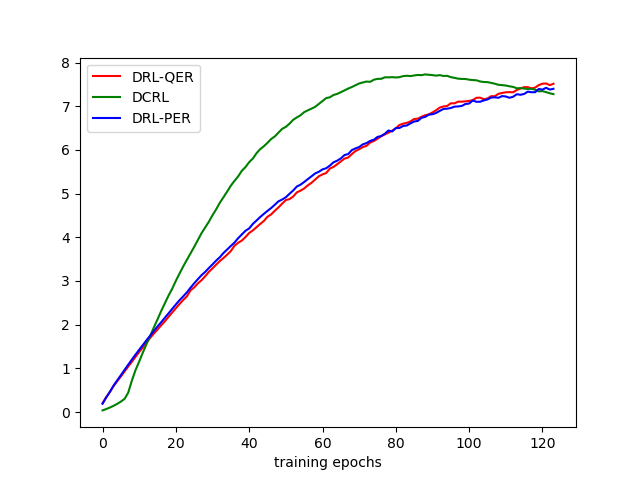}}
  \centerline{ (b) Carnival }
\end{minipage}
\hfill
\begin{minipage}{0.3\linewidth}
  \centerline{\includegraphics[width=2.4in,height=2in]{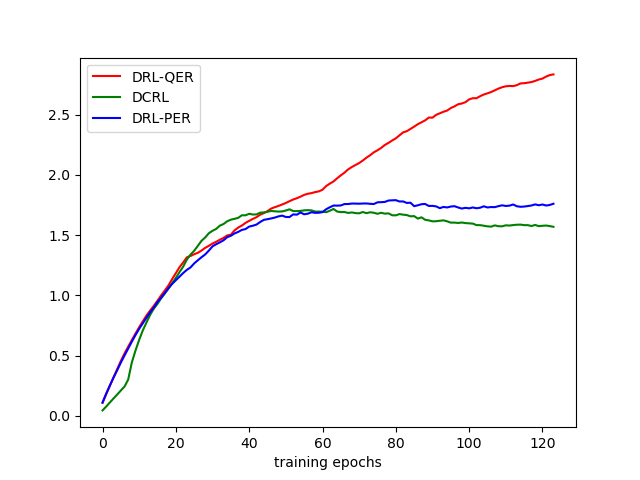}}
  \centerline{ (c) Breakout }
\end{minipage}
\vfill
\begin{minipage}{0.3\linewidth}
  \centerline{\includegraphics[width=2.4in,height=2in]{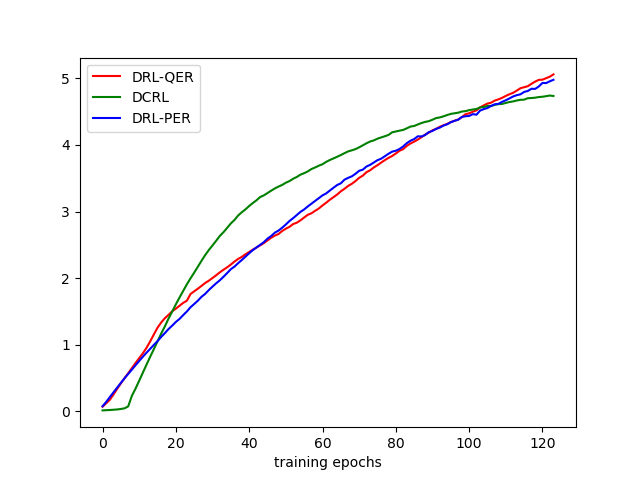}}
  \centerline{ (d) Freeway }
\end{minipage}
\hfill
\begin{minipage}{0.3\linewidth}
  \centerline{\includegraphics[width=2.4in,height=2in]{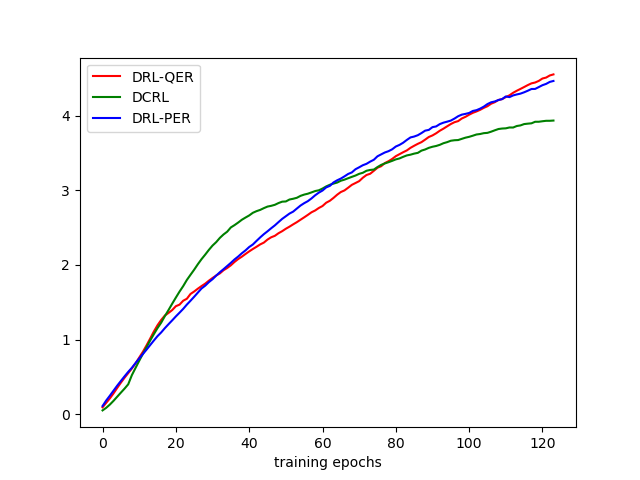}}
  \centerline{ (e) Beam Rider }
\end{minipage}
\hfill
\begin{minipage}{0.3\linewidth}
  \centerline{\includegraphics[width=2.4in,height=2in]{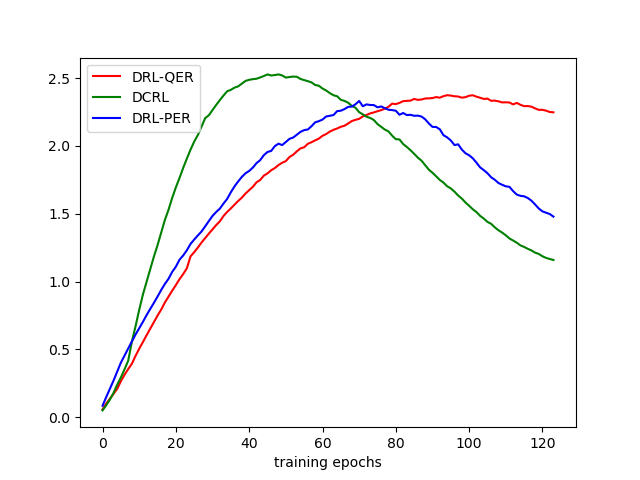}}
  \centerline{ (f) Kung-Fu Master }
\end{minipage}
\vfill
\begin{minipage}{0.3\linewidth}
  \centerline{\includegraphics[width=2.4in,height=2in]{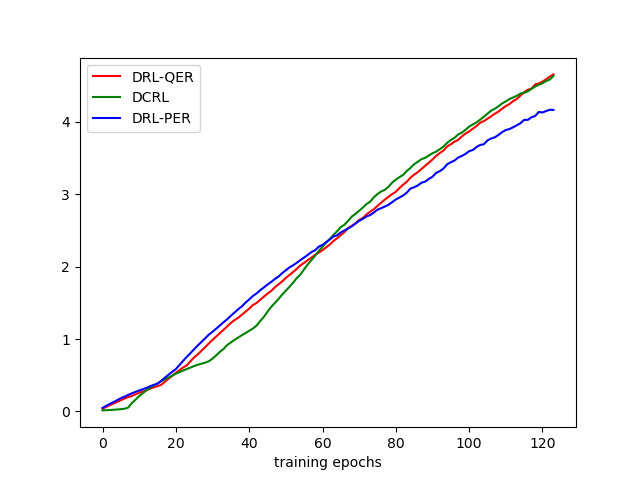}}
  \centerline{ (g) Road Runner }
\end{minipage}
\hfill
\begin{minipage}{0.3\linewidth}
  \centerline{\includegraphics[width=2.4in,height=2in]{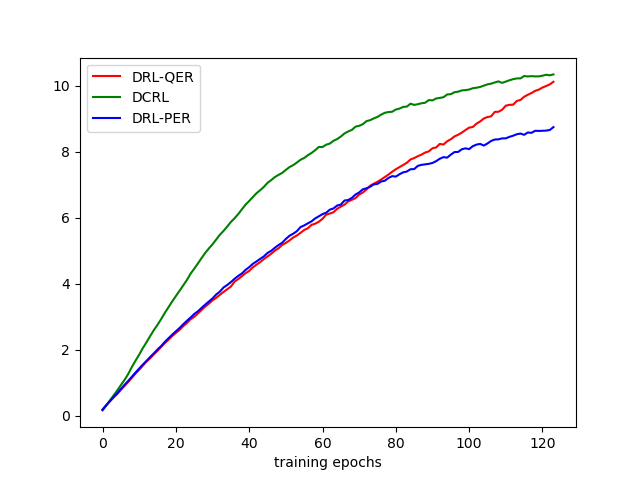}}
  \centerline{ (h) River Raid }
\end{minipage}
\hfill
\begin{minipage}{0.3\linewidth}
  \centerline{\includegraphics[width=2.4in,height=2in]{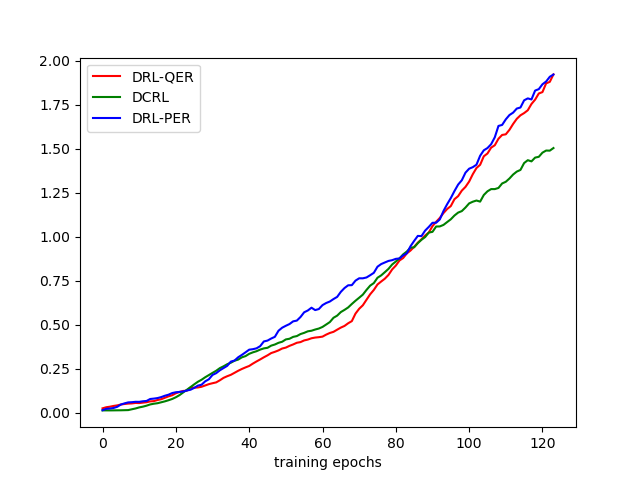}}
  \centerline{ (i) Enduro }
\end{minipage}
\vfill
\begin{minipage}{0.3\linewidth}
  \centerline{\includegraphics[width=2.4in,height=2in]{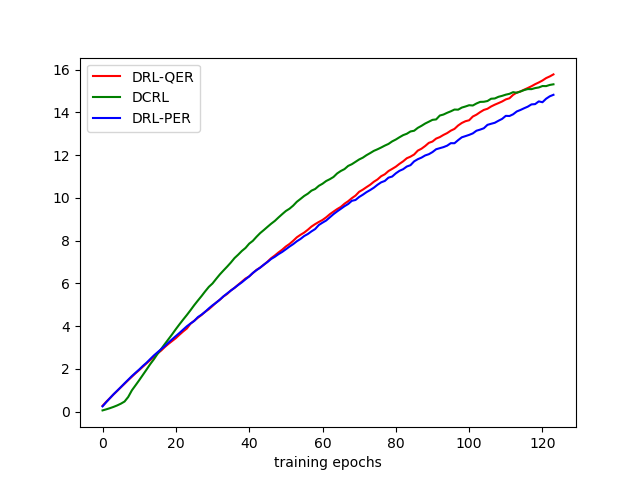}}
  \centerline{ (j) Ms. Pacman }
\end{minipage}
\hfill
\begin{minipage}{0.3\linewidth}
  \centerline{\includegraphics[width=2.4in,height=2in]{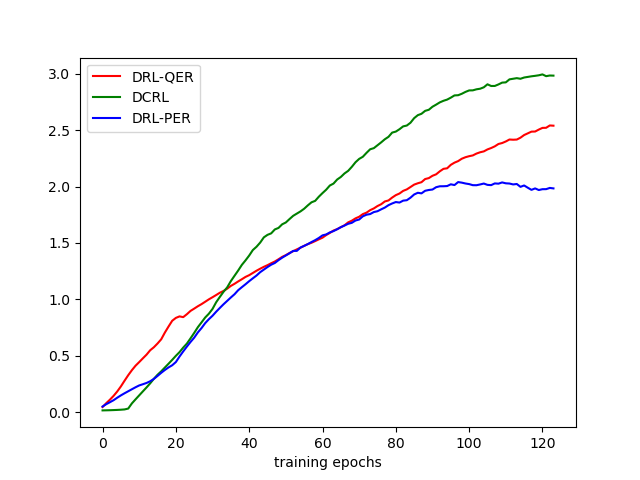}}
  \centerline{ (k) Kangaroo }
\end{minipage}
\hfill
\begin{minipage}{0.3\linewidth}
  \centerline{\includegraphics[width=2.4in,height=2in]{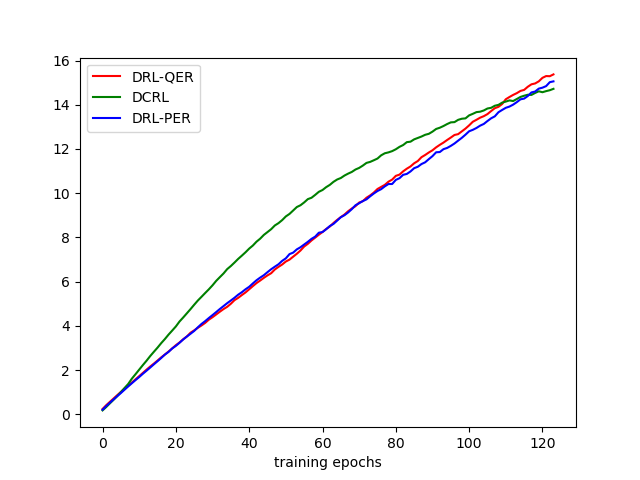}}
  \centerline{ (l) Alien }
\end{minipage}
%%%%%%%%%%%%%%%%%%%%%%%%%%%%%%%%%%%%%%%%%%%%%%%%%%%%%%%%%%%%%%%%%%%%%%
\caption{Performance of DRL-QER with comparison to  DRL-PER and DCRL
regarding the average Q value.} \label{Figure6}
\end{figure*}

\subsection{Experimental Results}\label{Sec4.2}
The experiments of the twelve games 
are deployed to compare the performance of DRL-PER, DCRL and DRL-QER.
Similar to DCRL \cite{ren2018}, AER \cite{SunZL20} and
CER \cite{LiuTSX19}, each simulation is run for three times to collect the average performance for a fair comparison. After the training process, we test the agents for 150 episodes, and the average rewards with the standard deviation are summarized in Table \ref{Table2}. It is clear that DRL-QER outperforms DRL-PER in most of the testing games. The statistical analysis also reveals that DRL-QER and DCRL achieve a comparative performance for the twelve games and they have different advantages in different games. 
Considering the total reward metric tends to be noisy because small changes to the weights of the DRL agent can lead to large changes in the distribution of states the agent visits \cite{mnih2015human}, we take the estimated action-value as the metric, which has been demonstrated to be more stable than reward metric to reveal the training performance of DRL methods. In particular,  we divide the training phase into 125 epochs, and the average action values of the testing frames are recorded after each training epoch.
 The experimental results demonstrate that the learning progress of DRL-QER is faster and more robust than that of DRL-PER, and is not worse than DCRL. It is worthy to note that DRL-QER merely changes the priorities of experiences without affecting the convergence of the baseline DRL method \cite{schaul2015prioritized}. However, an efficient use of samples contributes to a faster convergence under limited training epochs. Hence, the trends of the training curves (shown in Fig. \ref{Figure6}) reflect the superiority of our method.
What's more, DCRL involves many parameters which are difficult to fine tune for different games, while DRL-QER does not require prior knowledge to fine tune parameters. In fact, the parameter settings of DRL-QER are almost the same across the twelve games. From this respective, DRL-QER is an effective and general approach with enhanced performance.

%When it comes to comparing their experimental results,

%The experiment algorithms are based on those in \cite{van2016deep} and
%\cite{wang2015dueling}, and the hyper-parameters for these algorithms are the same as those in Table \ref{Table1}. We incorporate the quantum-inspired experience replay mechanism into the two DRL methods,

\subsection{Additional Exploratory Experiments}\label{Sec4.3}
\begin{figure*}[!htbp]
\begin{minipage}{0.2\linewidth}
  \centerline{\includegraphics[width=1.8in,height=1.6in]{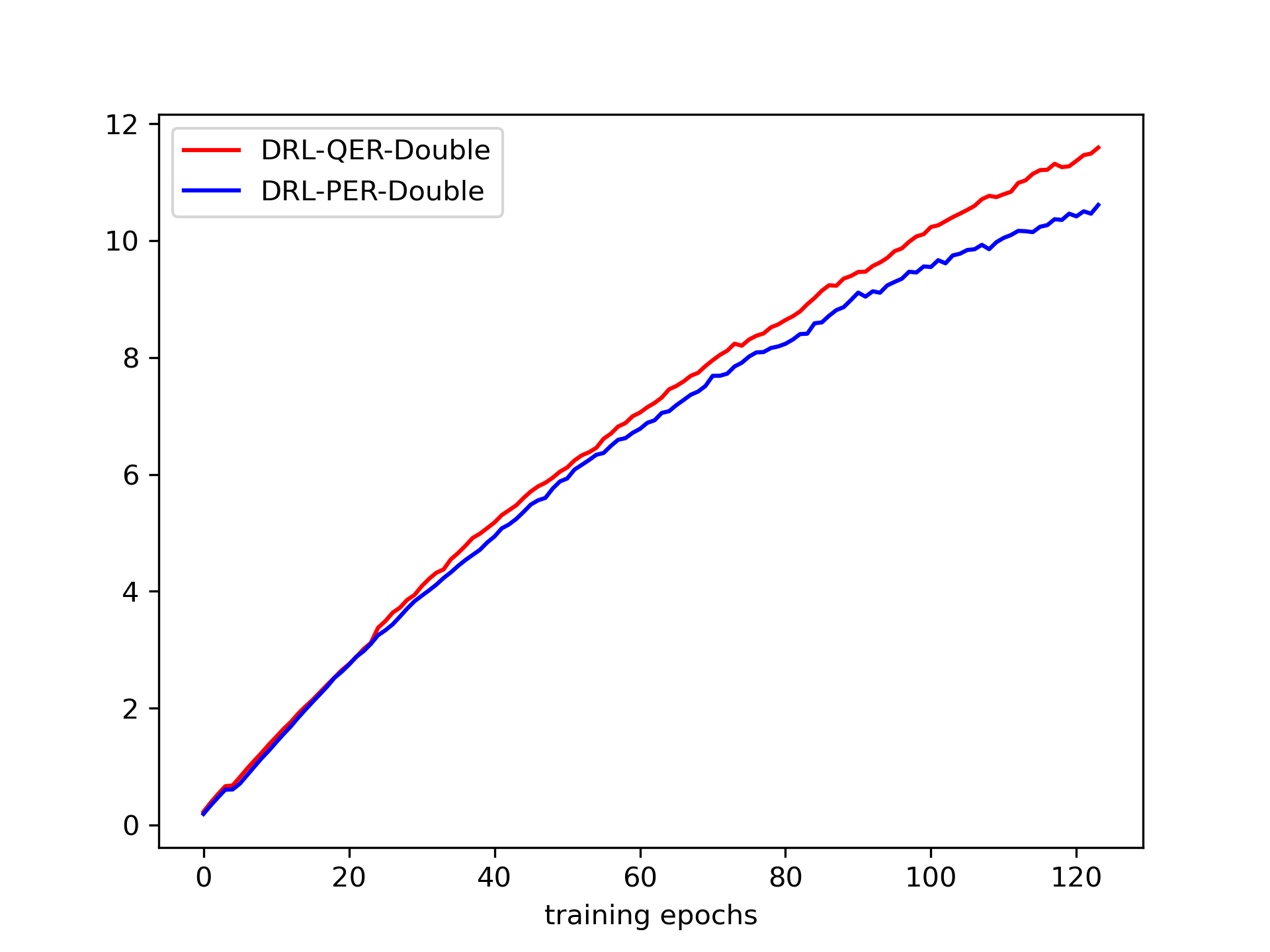}}
\end{minipage}
\hfill
\begin{minipage}{0.2\linewidth}
  \centerline{\includegraphics[width=1.8in,height=1.6in]{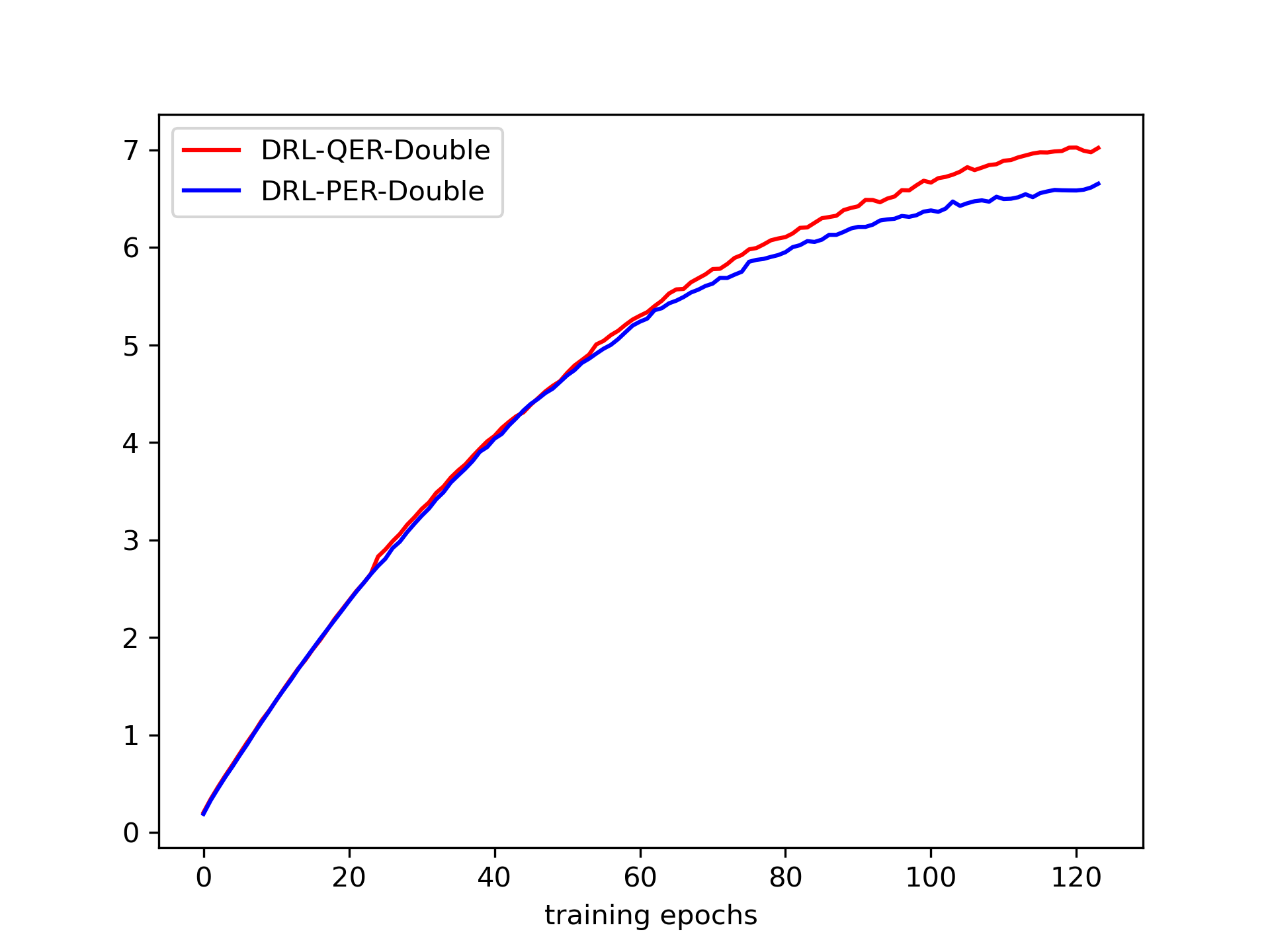}}
\end{minipage}
\hfill
\begin{minipage}{0.2\linewidth}
  \centerline{\includegraphics[width=1.8in,height=1.6in]{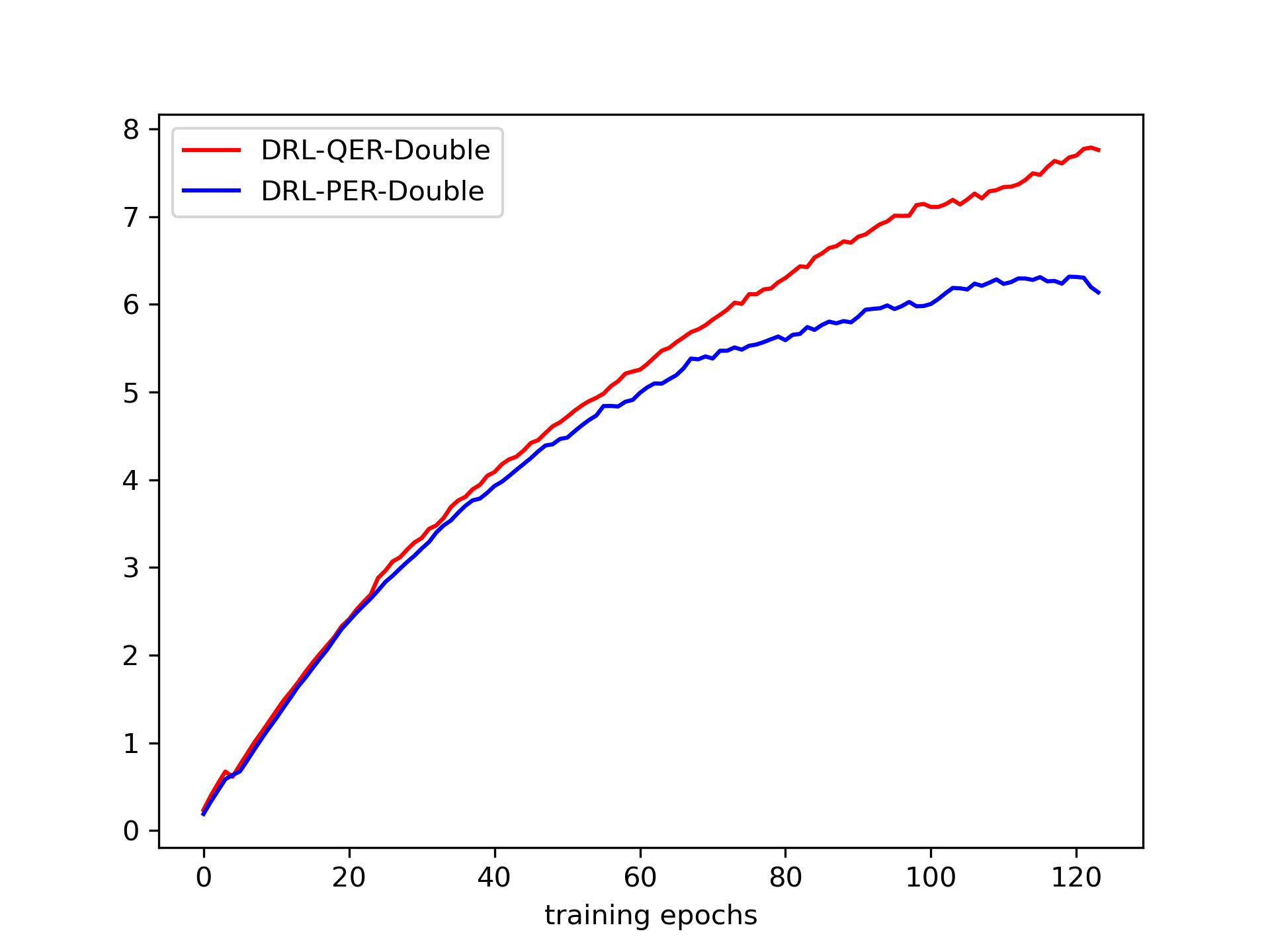}}
\end{minipage}
\hfill
\begin{minipage}{0.2\linewidth}
  \centerline{\includegraphics[width=1.8in,height=1.6in]{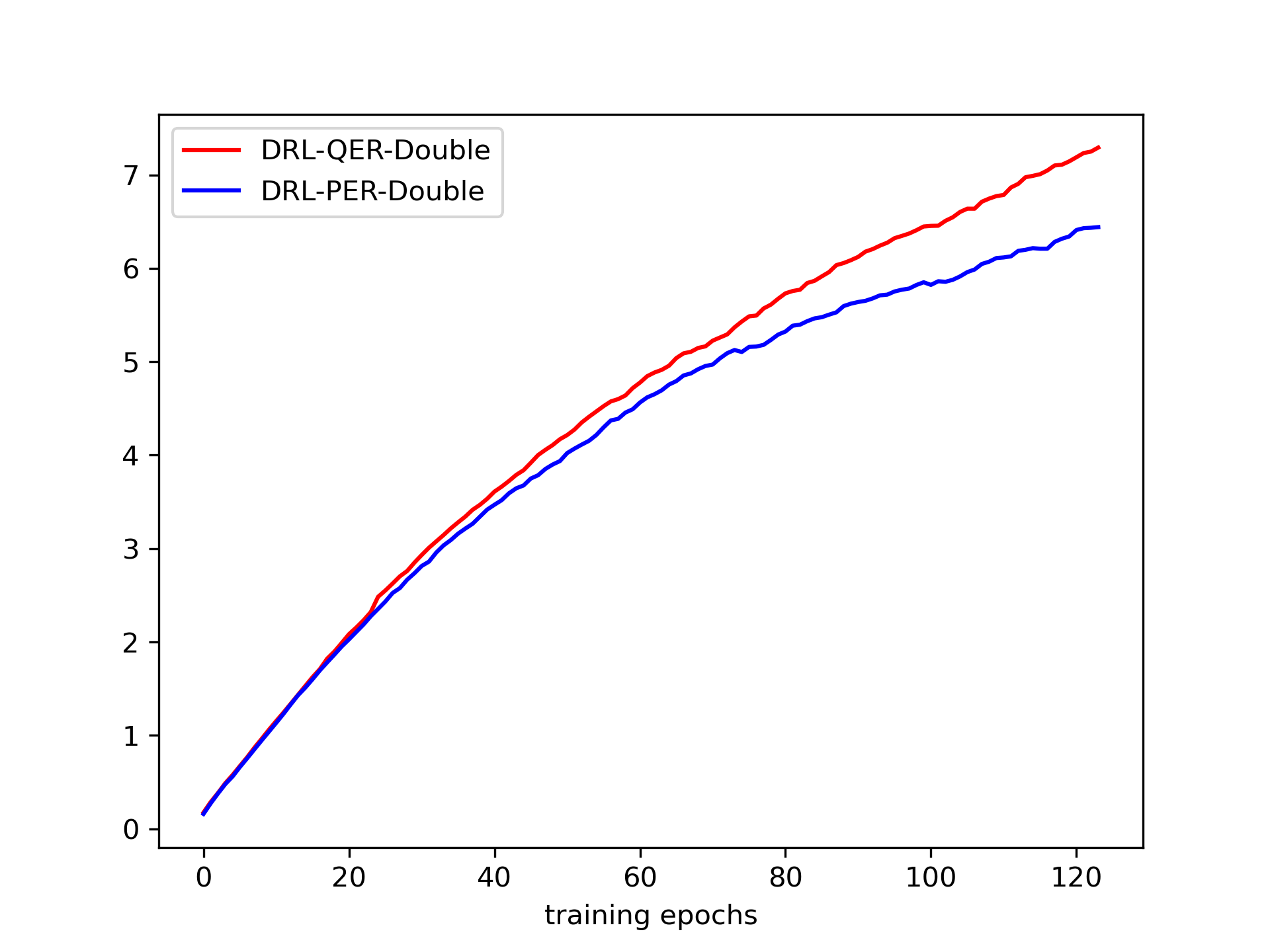}}
\end{minipage}
\vfill
\begin{minipage}{0.2\linewidth}
  \centerline{\includegraphics[width=1.8in,height=1.6in]{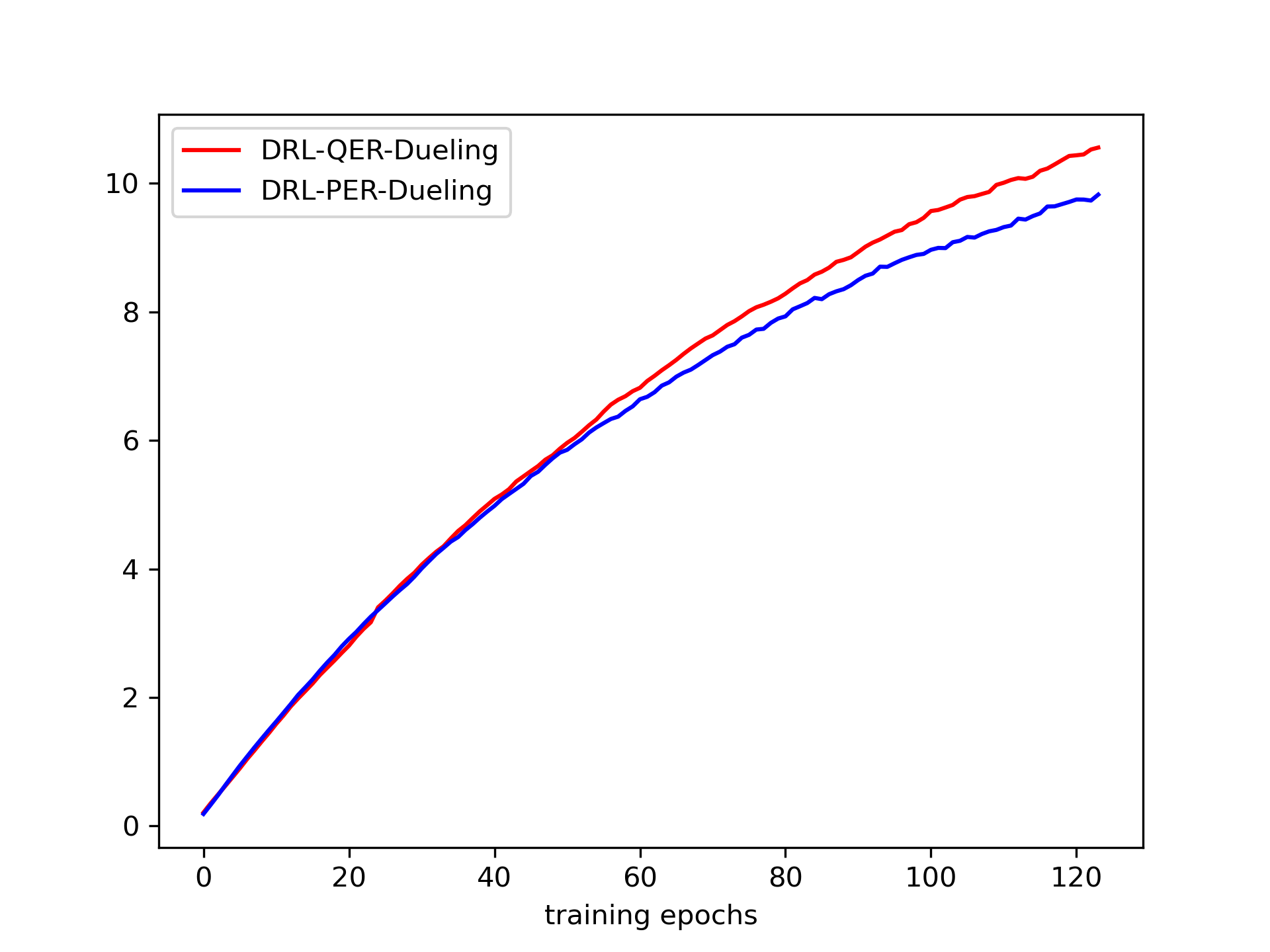}}
  \centerline{ (a) Alien }
\end{minipage}
\hfill
\begin{minipage}{0.2\linewidth}
  \centerline{\includegraphics[width=1.8in,height=1.6in]{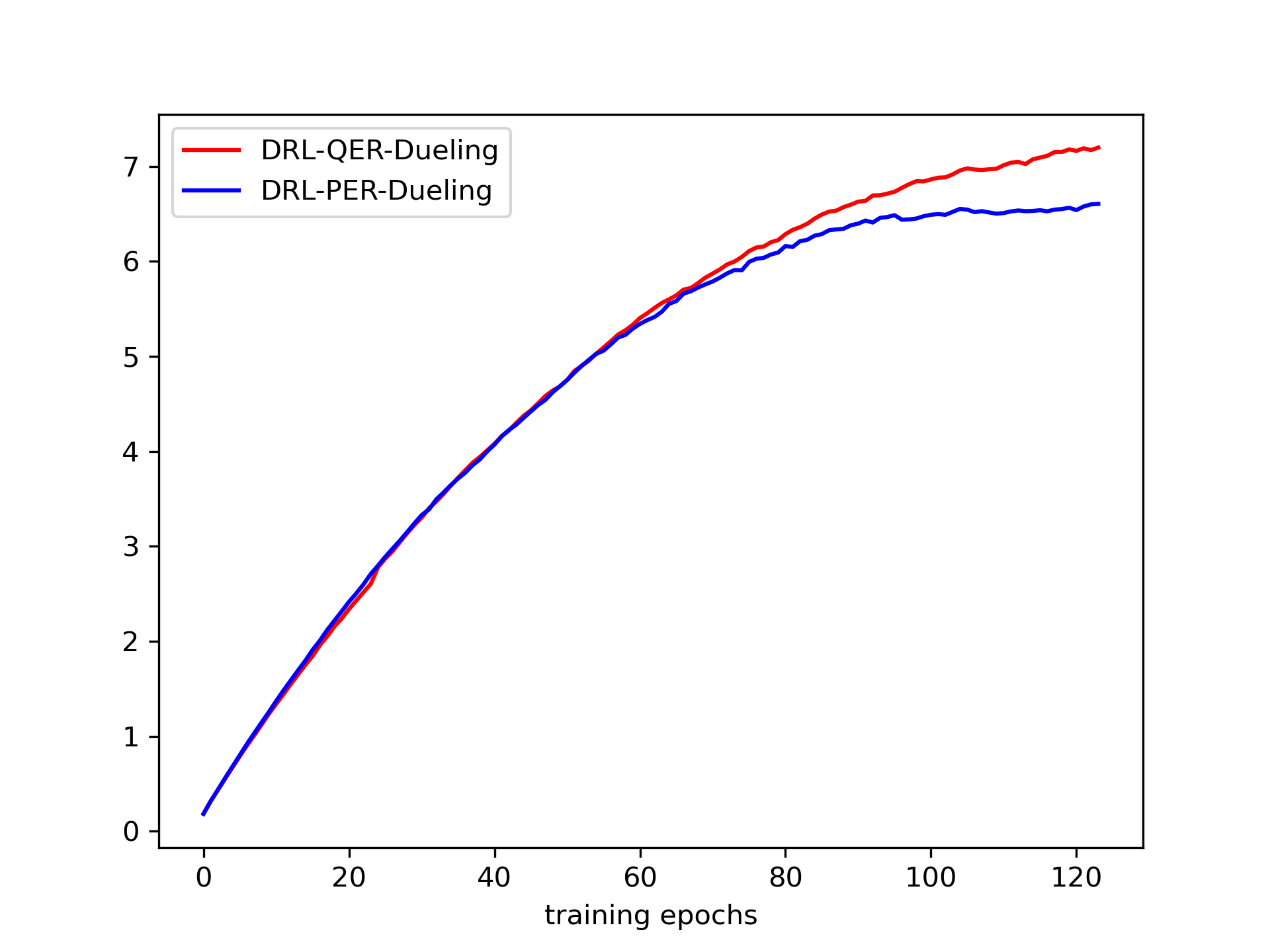}}
  \centerline{ (b) Carnival }
\end{minipage}
\hfill
\begin{minipage}{0.2\linewidth}
  \centerline{\includegraphics[width=1.8in,height=1.6in]{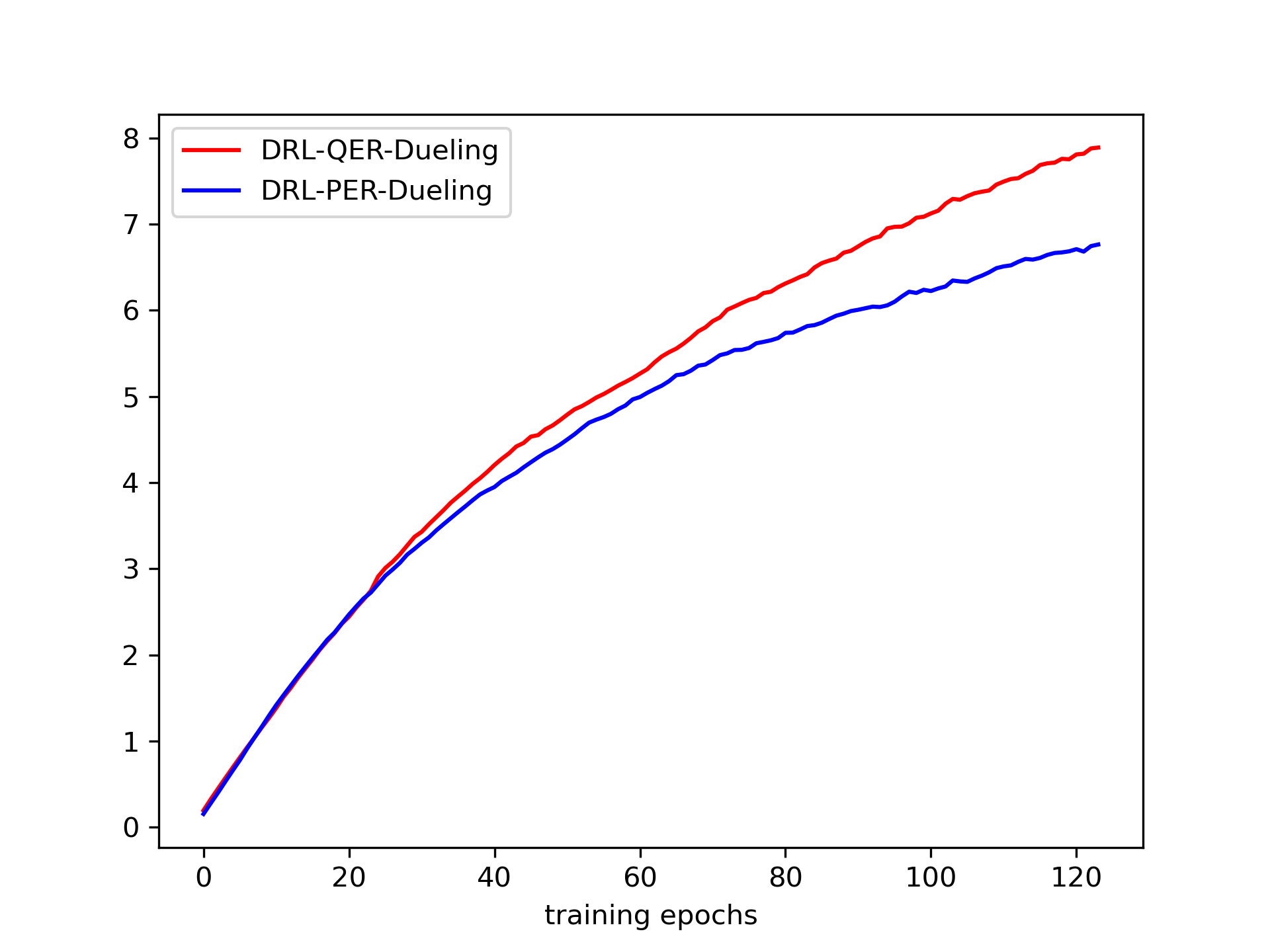}}
  \centerline{ (c) River Raid }
\end{minipage}
\hfill
\begin{minipage}{0.2\linewidth}
  \centerline{\includegraphics[width=1.8in,height=1.6in]{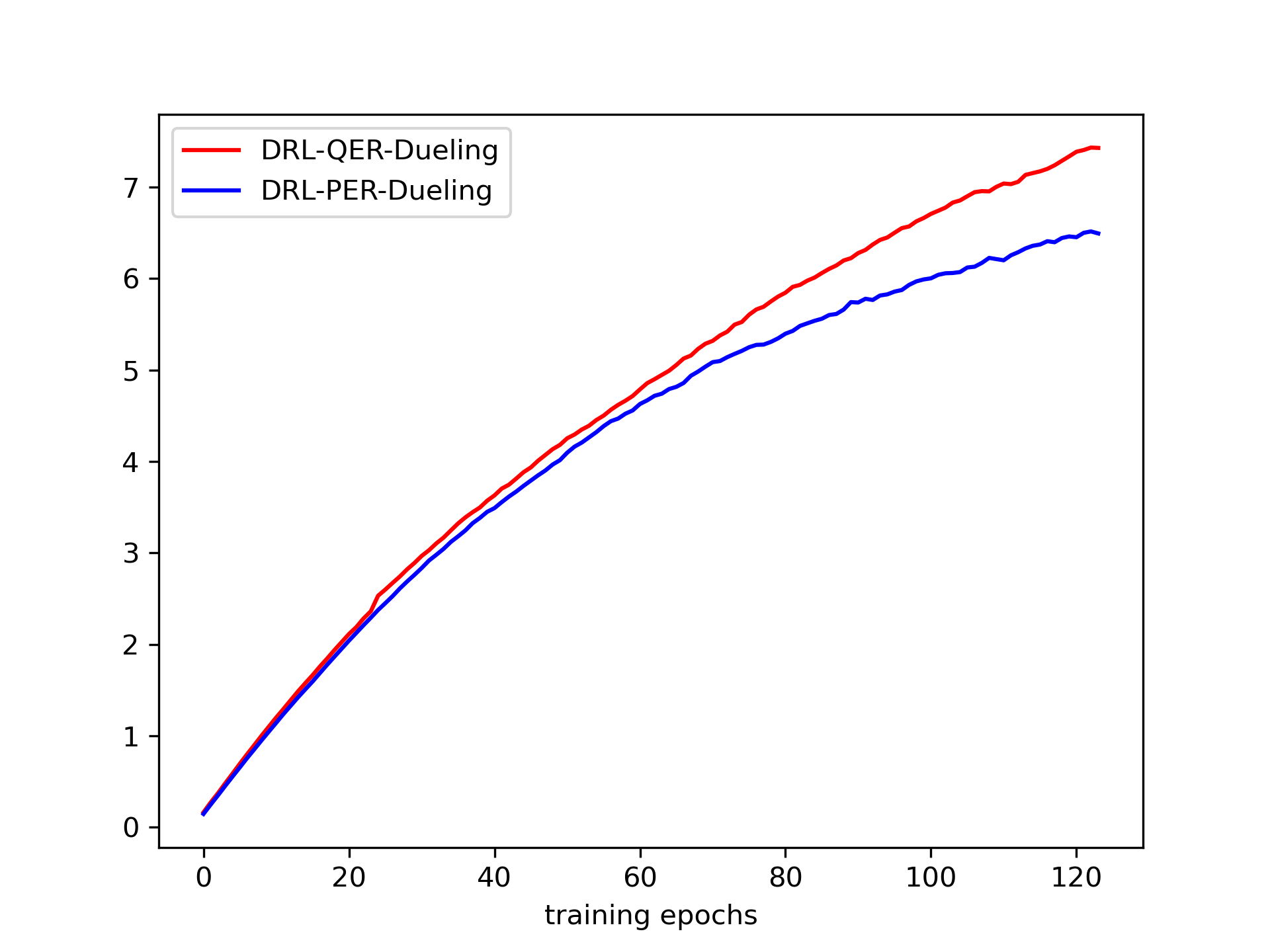}}
  \centerline{ (d) Space Invaders }
\end{minipage}
\caption{Performance of  DRL-QER-Double/DRL-PER-Double and DRL-QER-Dueling/DRL-PER-Dueling 
regarding the average Q value.} \label{Figure7}
\end{figure*}

The proposed DRL-QER aims at taking advantage of quantum characteristics in the experience replay mechanism. To figure out whether this mechanism can be applied to other memory-based RL algorithms, we further apply DRL-QER to double network
\cite{van2016deep} and dueling network \cite{wang2015dueling}, and implement experimental simulations on randomly selected 4 games using the same hyper-parameter setting in Table \ref{Table1}. In Fig. \ref{Figure7}, both double DQN and dueling DQN algorithms using the quantum-inspired experience replay (QER) method
show faster convergence regarding the average Q value compared with their classical counterparts. The average rewards of DRL-QER-Double and DRL-PER-Double are summarized in Table \ref{Table3}. Besides, the average rewards of DRL-QER-Dueling and DRL-PER-Dueling are summarized in Table \ref{Table4}. From these two tables, the average rewards per episode 
are also increased in "Double Network" and "Dueling Network" for 3 games except for Riverraid.

%The unitary operations determined by the amplitudes of TD-errors and the training steps is a major evaluation indicator for the experiences' values.

\section{Conclusion}\label{Sec5}
In this paper, the DRL-QER method is proposed by introducing quantum characteristics into
the process of experience replay in DRL to guarantee that the
learning scheme focuses on what the agent has learnt from the interaction with the environment
instead of the prior knowledge. In DRL-QER, the experiences are represented in quantum
states, whose amplitudes are correlated with the TD-errors and the replaying times.
In particular, the preparation operation and the depreciation operation in DRL-QER help
speed up the training progress and achieve an improved sampling efficiency. The experimental
results demonstrate superior performance of the proposed DRL-QER over  DRL-PER and DCRL.
Comparisons of DRL-PER and DRL-QER in dueling DQN and double DQN further show that DRL-QER can
also achieve improved performance for other memory-based
DRL algorithms. Our future work will focus on in-depth
theoretical research on the convergence of DRL-QER and quantum enhanced reinforcement learning along with
its applications to other continuous control methods such as deep
deterministic policy gradient (DDPG) \cite{lillicrap2015}, \cite{zhao2014} and Soft Actor Critic \cite{sac}.

%\section*{Acknowledgment}

%\begin{thebibliography}{10}
\bibliographystyle{ieeetr}
\bibliography{bibfile}

\begin{thebibliography}{10}

\bibitem{sutton2018reinforcement}
R.~S. Sutton and A.~G. Barto, {\em Reinforcement learning: An introduction (2nd
  Edition)}.
\newblock MIT Press, 2018.

\bibitem{Littman2015}
M.~L. Littman, ``Reinforcement learning improves behaviour from evaluative
  feedback,'' {\em Nature}, vol.~521, pp.~445--451, May 2015.

\bibitem{li2020quantum}
J.-A. Li, D.~Dong, Z.~Wei, Y.~Liu, Y.~Pan, F.~Nori, and X.~Zhang, ``Quantum
  reinforcement learning during human decision-making,'' {\em Nature Human
  Behaviour}, vol.~4, pp.~294--307, March 2020.

\bibitem{krizhevsky2012imagenet}
A.~Krizhevsky, I.~Sutskever, and G.~E. Hinton, ``Imagenet classification with
  deep convolutional neural networks,'' in {\em Advances in Neural Information
  Processing Systems}, pp.~1097--1105, 2012.

\bibitem{farabet2012learning}
C.~{Farabet}, C.~{Couprie}, L.~{Najman}, and Y.~{LeCun}, ``Learning
  hierarchical features for scene labeling,'' {\em IEEE Transactions on Pattern
  Analysis and Machine Intelligence}, vol.~35, pp.~1915--1929, August 2013.

\bibitem{goh2014learning}
H.~Goh, N.~Thome, M.~Cord, and J.-H. Lim, ``Learning deep hierarchical visual
  feature coding,'' {\em IEEE Transactions on Neural Networks and Learning
  Systems}, vol.~25, pp.~2212--2225, December 2014.

\bibitem{hinton2012deep}
H.~Geoffrey, D.~Li, Y.~Dong, E.~D. George, and A.-r. Mohamed, ``Deep neural
  networks for acoustic modeling in speech recognition: The shared views of
  four research groups,'' {\em IEEE Signal Processing Magazine}, vol.~29,
  pp.~82--97, November 2012.

\bibitem{sutskever2014sequence}
I.~Sutskever, O.~Vinyals, and Q.~V. Le, ``Sequence to sequence learning with
  neural networks,'' in {\em Advances in Neural Information Processing
  Systems}, pp.~3104--3112, 2014.

\bibitem{brahma2015deep}
P.~P. Brahma, D.~Wu, and Y.~She, ``Why deep learning works: A manifold
  disentanglement perspective,'' {\em IEEE Transactions on Neural Networks and
  Learning Systems}, vol.~27, pp.~1997--2008, October 2015.

\bibitem{Tembine2020}
H.~{Tembine}, ``Deep learning meets game theory: Bregman-based algorithms for
  interactive deep generative adversarial networks,'' {\em IEEE Transactions on
  Cybernetics}, vol.~50, pp.~1132--1145, March 2020.

\bibitem{mnih2015human}
V.~Mnih, K.~Kavukcuoglu, D.~Silver, A.~A. Rusu, J.~Veness, M.~G. Bellemare,
  A.~Graves, M.~Riedmiller, A.~K. Fidjeland, G.~Ostrovski, {\em et~al.},
  ``Human-level control through deep reinforcement learning,'' {\em Nature},
  vol.~518, p.~529, February 2015.

\bibitem{lin1992self}
L.-J. Lin, ``Self-improving reactive agents based on reinforcement learning,
  planning and teaching,'' {\em Machine Learning}, vol.~8, no.~3-4,
  pp.~293--321, 1992.

\bibitem{schaul2015prioritized}
T.~Schaul, J.~Quan, I.~Antonoglou, and D.~Silver, ``Prioritized experience
  replay,'' in {\em Proceedings of the IEEE International Conference on
  Learning Representations}, 2016.

\bibitem{ren2018}
Z.~Ren, D.~Dong, H.~Li, and C.~Chen, ``Self-paced prioritized curriculum
  learning with coverage penalty in deep reinforcement learning,'' {\em IEEE
  Transactions on Neural Networks and Learning Systems}, vol.~29,
  pp.~2216--2226, June 2018.

\bibitem{pmlr-v97-novati19a}
G.~Novati and P.~Koumoutsakos, ``Remember and forget for experience replay,''
  in {\em Proceedings of the 36th International Conference on Machine Learning}
  (K.~Chaudhuri and R.~Salakhutdinov, eds.), vol.~97 of {\em Proceedings of
  Machine Learning Research}, (Long Beach, California, USA), pp.~4851--4860,
  PMLR, 09--15 Jun 2019.

\bibitem{SunZL20}
P.~Sun, W.~Zhou, and H.~Li, ``Attentive experience replay,'' in {\em The
  Thirty-Fourth {AAAI} Conference on Artificial Intelligence, {AAAI} 2020, The
  Thirty-Second Innovative Applications of Artificial Intelligence Conference,
  {IAAI} 2020, The Tenth {AAAI} Symposium on Educational Advances in Artificial
  Intelligence, {EAAI} 2020, New York, NY, USA, February 7-12, 2020},
  pp.~5900--5907, {AAAI} Press, 2020.

\bibitem{LiuTSX19}
H.~Liu, A.~Trott, R.~Socher, and C.~Xiong, ``Competitive experience replay,''
  in {\em 7th International Conference on Learning Representations, {ICLR}
  2019, New Orleans, LA, USA, May 6-9, 2019}, OpenReview.net, 2019.

\bibitem{chang2017active}
H.-S. Chang, E.~Learned-Miller, and A.~McCallum, ``Active bias: Training more
  accurate neural networks by emphasizing high variance samples,'' in {\em
  Advances in Neural Information Processing Systems}, pp.~1002--1012, 2017.

\bibitem{shor1994algorithms}
P.~W. Shor, ``Algorithms for quantum computation: Discrete logarithms and
  factoring,'' in {\em Proceedings 35th Annual Symposium on Foundations of
  Computer Science}, pp.~124--134, 1994.

\bibitem{grover1997}
L.~K. Grover, ``Quantum computers can search arbitrarily large databases by a
  single query,'' {\em Physical Review Letters}, vol.~79, pp.~4709--4712,
  December 1997.

\bibitem{Biamonte2017}
J.~Biamonte, P.~Wittek, N.~Pancotti, P.~Rebentrost, N.~Wiebe, and S.~Lloyd,
  ``Quantum machine learning,'' {\em Nature}, vol.~549, pp.~195--202, September
  2017.

\bibitem{Cai2015}
X.-D. Cai, D.~Wu, Z.-E. Su, M.-C. Chen, X.-L. Wang, L.~Li, N.-L. Liu, C.-Y. Lu,
  and J.-W. Pan, ``Entanglement-based machine learning on a quantum computer,''
  {\em Physical Review Letters}, vol.~114, p.~110504, March 2015.

\bibitem{Li2015}
Z.~Li, X.~Liu, N.~Xu, and J.~Du, ``Experimental realization of a quantum
  support vector machine,'' {\em Physical Review Letters}, vol.~114, p.~140504,
  April 2015.

\bibitem{Beer2020}
K.~Beer, D.~Bondarenko, T.~Farrelly, T.~J. Osborne, R.~Salzmann,
  D.~Scheiermann, and R.~Wolf, ``Training deep quantum neural networks,'' {\em
  Nature Communications}, vol.~11, p.~808, 2020.

\bibitem{Bai2020}
L.~{Bai}, L.~{Rossi}, L.~{Cui}, J.~{Cheng}, and E.~R. {Hancock}, ``A
  quantum-inspired similarity measure for the analysis of complete weighted
  graphs,'' {\em IEEE Transactions on Cybernetics}, vol.~50, pp.~1264--1277,
  March 2020.

\bibitem{Ding2019}
W.~{Ding}, C.~{Lin}, and Z.~{Cao}, ``Deep neuro-cognitive co-evolution for
  fuzzy attribute reduction by quantum leaping {PSO} with nearest-neighbor
  memeplexes,'' {\em IEEE Transactions on Cybernetics}, vol.~49,
  pp.~2744--2757, July 2019.

\bibitem{kak1995quantum}
S.~C. Kak, ``Quantum neural computing,'' in {\em Advances in Imaging and
  Electron Physics}, vol.~94, pp.~259--313, Elsevier, 1995.

\bibitem{dunjko2018machine}
V.~Dunjko and H.~J. Briegel, ``Machine learning \& artificial intelligence in
  the quantum domain: a review of recent progress,'' {\em Reports on Progress
  in Physics}, no.~7, p.~074001, 2018.

\bibitem{Carleo602}
G.~Carleo and M.~Troyer, ``Solving the quantum many-body problem with
  artificial neural networks,'' {\em Science}, vol.~355, pp.~602--606, February
  2017.

\bibitem{Lloyd2014}
S.~Lloyd, M.~Mohseni, and P.~Rebentrost, ``Quantum principal component
  analysis,'' {\em Nature Physics}, vol.~10, pp.~631--633, September 2014.

\bibitem{Dong2008}
D.~Dong, C.~Chen, H.~Li, and T.~J. Tarn, ``Quantum reinforcement learning,''
  {\em {IEEE} Trans. Systems, Man, and Cybernetics, Part {B}}, vol.~38,
  pp.~1207--1220, October 2008.

\bibitem{paparo2014quantum}
G.~D. Paparo, V.~Dunjko, A.~Makmal, M.~A. Martin-Delgado, and H.~J. Briegel,
  ``Quantum speedup for active learning agents,'' {\em Physical Review X},
  vol.~4, no.~3, p.~031002, 2014.

\bibitem{Dunjko2016}
V.~Dunjko, J.~M. Taylor, and H.~J. Briegel, ``Quantum-enhanced machine
  learning,'' {\em Physical Review Letters}, vol.~117, p.~130501, September
  2016.

\bibitem{Lamata2017}
L.~Lamata, ``Basic protocols in quantum reinforcement learning with
  superconducting circuits,'' {\em Scientific Reports}, vol.~7, p.~1609, May
  2017.

\bibitem{cardenas2018multiqubit}
F.~C{\'a}rdenas-L{\'o}pez, L.~Lamata, J.~Retamal, and E.~Solano, ``Multiqubit
  and multilevel quantum reinforcement learning with quantum technologies,''
  {\em PloS One}, vol.~13, no.~7, p.~e0200455, 2018.

\bibitem{hu2019training}
W.~Hu and J.~Hu, ``Training a quantum neural network to solve the contextual
  multi-armed bandit problem,'' {\em Natural Science}, vol.~11, no.~1,
  pp.~17--27, 2019.

\bibitem{crawford2016reinforcement}
D.~Crawford, A.~Levit, N.~Ghadermarzy, J.~S. Oberoi, and P.~Ronagh,
  ``Reinforcement learning using quantum boltzmann machines,'' {\em Quantum
  Information {\&} Computation}, vol.~18, no.~1{\&}2, pp.~51--74, 2018.

\bibitem{nielsen2010quantum}
M.~A. Nielsen and I.~L. Chuang, {\em Quantum Computation and Quantum
  Information}.
\newblock Cambridge University Press, 2010.

\bibitem{baird1995residual}
L.~Baird, ``Residual algorithms: Reinforcement learning with function
  approximation,'' in {\em Machine Learning Proceedings 1995}, pp.~30--37,
  Elsevier, 1995.

\bibitem{Luo2018}
B.~{Luo}, Y.~{Yang}, and D.~{Liu}, ``Adaptive {$Q$}-learning for data-based
  optimal output regulation with experience replay,'' {\em IEEE Transactions on
  Cybernetics}, vol.~48, pp.~3337--3348, December 2018.

\bibitem{Ni2019}
Z.~{Ni}, N.~{Malla}, and X.~{Zhong}, ``Prioritizing useful experience replay
  for heuristic dynamic programming-based learning systems,'' {\em IEEE
  Transactions on Cybernetics}, vol.~49, pp.~3911--3922, November 2019.

\bibitem{dong2010quantum}
D.~Dong and I.~R. Petersen, ``Quantum control theory and applications: a
  survey,'' {\em IET Control Theory \& Applications}, vol.~4, no.~12,
  pp.~2651--2671, 2010.

\bibitem{de2015importance}
T.~De~Bruin, J.~Kober, K.~Tuyls, and R.~Babu{\v{s}}ka, ``The importance of
  experience replay database composition in deep reinforcement learning,'' in
  {\em Deep Reinforcement Learning Workshop, NIPS}, 2015.

\bibitem{ishii2002control}
S.~Ishii, W.~Yoshida, and J.~Yoshimoto, ``Control of exploitation--exploration
  meta-parameter in reinforcement learning,'' {\em Neural Networks}, vol.~15,
  no.~4-6, pp.~665--687, 2002.

\bibitem{abbeel2005exploration}
P.~Abbeel and A.~Y. Ng, ``Exploration and apprenticeship learning in
  reinforcement learning,'' in {\em Proceedings of the 22nd International
  Conference on Machine Learning}, pp.~1--8, ACM, 2005.

\bibitem{mannucci2017safe}
T.~Mannucci, E.-J. van Kampen, C.~de~Visser, and Q.~Chu, ``Safe exploration
  algorithms for reinforcement learning controllers,'' {\em IEEE Transactions
  on Neural Networks and Learning Systems}, vol.~29, pp.~1069--1081, April
  2017.

\bibitem{Dong2012robust}
D.~{Dong}, C.~{Chen}, J.~{Chu}, and T.~{Tarn}, ``Robust quantum-inspired
  reinforcement learning for robot navigation,'' {\em IEEE/ASME Transactions on
  Mechatronics}, vol.~17, pp.~86--97, February 2012.

\bibitem{Chen2014Fidelity}
C.~{Chen}, D.~{Dong}, H.~{Li}, J.~{Chu}, and T.~{Tarn}, ``Fidelity-based
  probabilistic q-learning for control of quantum systems,'' {\em IEEE
  Transactions on Neural Networks and Learning Systems}, vol.~25, pp.~920--933,
  May 2014.

\bibitem{Zhang2019}
Q.~{Zhang} and D.~{Zhao}, ``Data-based reinforcement learning for nonzero-sum
  games with unknown drift dynamics,'' {\em IEEE Transactions on Cybernetics},
  vol.~49, pp.~2874--2885, August 2019.

\bibitem{Gym2016}
G.~Brockman, V.~Cheung, L.~Pettersson, J.~Schneider, J.~Schulman, J.~Tang, and
  W.~Zaremba, ``Openai gym,'' {\em CoRR}, vol.~abs/1606.01540, 2016.

\bibitem{van2016deep}
H.~van Hasselt, A.~Guez, and D.~Silver, ``Deep reinforcement learning with
  double {Q}-learning,'' in {\em Proceedings of the Thirtieth {AAAI} Conference
  on Artificial Intelligence, Phoenix, Arizona, {USA}, February 12-17, 2016},
  pp.~2094--2100.

\bibitem{wang2015dueling}
Z.~Wang, T.~Schaul, M.~Hessel, H.~van Hasselt, M.~Lanctot, and N.~de~Freitas,
  ``Dueling network architectures for deep reinforcement learning,'' in {\em
  Proceedings of the 33nd International Conference on Machine Learning, New
  York, USA, June 19-24, 2016}, pp.~1995--2003.

\bibitem{lillicrap2015}
T.~P. Lillicrap, J.~J. Hunt, A.~Pritzel, N.~Heess, T.~Erez, Y.~Tassa,
  D.~Silver, and D.~Wierstra, ``Continuous control with deep reinforcement
  learning,'' in {\em 4th International Conference on Learning
  Representations}, 2016.

\bibitem{zhao2014}
D.~Zhao and Y.~Zhu, ``Mec—a near-optimal online reinforcement learning
  algorithm for continuous deterministic systems,'' {\em IEEE Transactions on
  Neural Networks and Learning Systems}, vol.~26, no.~2, pp.~346--356, 2014.

\bibitem{sac}
T.~Haarnoja, A.~Zhou, P.~Abbeel, and S.~Levine, ``Soft actor-critic: Off-policy
  maximum entropy deep reinforcement learning with a stochastic actor,'' in
  {\em Proceedings of the 35th International Conference on Machine Learning,
  {ICML} 2018, Stockholmsm{\"{a}}ssan, Stockholm, Sweden, July 10-15, 2018}
  (J.~G. Dy and A.~Krause, eds.), vol.~80 of {\em Proceedings of Machine
  Learning Research}, pp.~1856--1865, {PMLR}, 2018.

\end{thebibliography}

\end{document}